\documentclass[10pt,journal,compsoc]{IEEEtran}

\ifCLASSOPTIONcompsoc
  \usepackage[nocompress]{cite}
\else
  \usepackage{cite}
\fi

\usepackage{graphicx}
\usepackage{amsmath}
\usepackage{amssymb}
\usepackage{booktabs}
\usepackage{makecell}
\usepackage{adjustbox}
\usepackage{siunitx}
\usepackage{multirow}

\newcolumntype{L}{>{\let\newline\\\arraybackslash\hspace{0pt}\raggedright}m{3cm}}
\newcolumntype{T}{>{\let\newline\\\arraybackslash\hspace{0pt}\raggedright}m{2.2cm}}

\usepackage{stfloats}

\usepackage[table, dvipsnames]{xcolor}
\usepackage{wrapfig}

\usepackage{tikz}
\usetikzlibrary{backgrounds}
\usetikzlibrary{arrows,shapes}
\usetikzlibrary{tikzmark}
\usetikzlibrary{calc}
\newcommand{\highlight}[2]{\colorbox{#1!17}{$\displaystyle #2$}}

\renewcommand{\highlight}[2]{\colorbox{#1!17}{#2}}

\usepackage[breaklinks,colorlinks]{hyperref}
\usepackage{float}
\usepackage{enumitem}

\usepackage[capitalize]{cleveref}
\crefname{section}{Sec.}{Secs.}
\Crefname{section}{Section}{Sections}
\Crefname{table}{Table}{Tables}
\crefname{table}{Tab.}{Tabs.}

\makeatletter
\newcommand{\ssymbol}[1]{^{\@fnsymbol{#1}}}
\makeatother

\begin{document}

\title{A Comparative Attention Framework for Better Few-Shot Object Detection on Aerial Images}

\author{Pierre~Le~Jeune and Anissa~Mokraoui
\IEEEcompsocitemizethanks{
\IEEEcompsocthanksitem Pierre Le Jeune is with L2TI, Université Sorbonne Paris Nord and COSE.
\IEEEcompsocthanksitem Anissa Mokraoui is with L2TI, Université Sorbonne Paris Nord.
\IEEEcompsocthanksitem Corresponding E-mail: pierre.le-jeune@cose.fr}
}

\IEEEtitleabstractindextext{%
\begin{abstract}
  Few-Shot Object Detection (FSOD) methods are mainly designed and evaluated on
  natural image datasets such as Pascal VOC and MS COCO. However, it is not
  clear whether the best methods for natural images are also the best for aerial
  images. Furthermore, direct comparison of performance between FSOD methods is
  difficult due to the wide variety of detection frameworks and training
  strategies. Therefore, we propose a benchmarking framework that provides a
  flexible environment to implement and compare attention-based FSOD methods.
  The proposed framework focuses on attention mechanisms and is divided into
  three modules: spatial alignment, global attention, and fusion layer. To
  remain competitive with existing methods, which often leverage complex
  training, we propose new augmentation techniques designed for object
  detection. Using this framework, several FSOD methods are reimplemented and
  compared. This comparison highlights two distinct performance regimes on
  aerial and natural images: FSOD performs worse on aerial images. Our
  experiments suggest that small objects, which are harder to detect in the
  few-shot setting, account for the poor performance. Finally, we develop a
  novel multiscale alignment method, Cross-Scales Query-Support Alignment (XQSA)
  for FSOD, to improve the detection of small objects. XQSA outperforms the
  state-of-the-art significantly on DOTA and DIOR.

\end{abstract}

}

\maketitle

\IEEEdisplaynontitleabstractindextext

\IEEEpeerreviewmaketitle

\IEEEraisesectionheading{\section{Introduction}\label{sec:introduction}}
\IEEEPARstart{I}{mpressive} progress has been made in object detection in the
past decade, mostly because of deep convolutional networks (see e.g.
\cite{ren2015faster,redmon2016you}). Current state-of-the-art performs
high-quality object detection, but it requires large annotated datasets and days
of training to achieve such quality. Often, these requirements could not be met,
and it is quite hard to achieve good performance for a specific application.
Few-shot learning focuses specifically on this kind of use cases where data is
scarce. It has been extensively studied for classification in the past years.
However, detection is a more challenging task and has been only tackled recently
from a few-shot perspective.

Few-Shot Object Detection (FSOD) is a challenging problem that aims to find all
occurrences of a class in an image given only a few examples. Current FSOD
state-of-the-art is mainly based on attention mechanisms, which aim at
extracting information about the task (i.e. semantic features about the classes
to be detected) from \textit{support} examples. Hence, the network can condition
the detection on the examples. A seminal work in this direction is presented in
reference \cite{kang2019few} which reweights features from the input image (also
called \textit{query} image) with the features extracted from the support
images. Plenty of methods based on the same idea have since been introduced (see
e.g. \cite{deng2020few,fan2020few,wallach2019one,li2020one}).
Although the attention mechanisms proposed in these papers differ
from the original one, the main principle remains
the same. It combines information from the query image and the support
examples to detect only the objects annotated in the support set.

The FSOD field is rapidly growing, and most new papers propose a novel attention
technique. However, there are a lot of design choices that can be considered to
address the FSOD problem. The detection framework and its backbone, the loss
function and all the hyperparameters that are tied to these methods make the
comparison between FSOD methods difficult. This paper focuses first on filling
this shortcoming. To do so, we propose a modular attention framework that
regroups most attention-based methods under the same notations. Specifically,
this framework is divided into three parts: spatial alignment, global attention,
and fusion layer. This separation helps to easily implement the different
mechanisms and facilitates the comparison. Most importantly, this makes it
possible to completely fix the other parameters and design choices without
reimplementing each method from scratch. That way, a fair comparison of
different attention mechanisms is possible. To help the development of future
FSOD comparisons, the code of the proposed framework is available
\footnote{\url{https://github.com/pierlj/aaf_framework}}. Some FSOD methods are
selected and reimplemented inside the framework to demonstrate its efficacy.
Most of these methods have originally been tested on Pascal VOC
\cite{everingham2010pascal} and MS COCO \cite{lin2014microsoft}. However, when
it comes to others datasets, specifically aerial images, there lacks comparative
studies about FSOD.
Therefore, we propose a comparison of several competitive FSOD methods on two
aerial datasets: DOTA \cite{xia2018dota} and DIOR \cite{li2020object}. It
appears that FSOD methods perform poorly on aerial images, in comparison with
classical detection (i.e. without few-shot). We hypothesize that this is due to
a greater proportion of small objects in aerial images. They are notoriously
hard to detect, even when plenty of examples are available. In FSOD, it is even
more difficult. Not only the model needs to detect small objects, but also it
needs to figure out what class to detect from small examples. It makes the small
FSOD much harder than small objects detection as the error can also come from
the conditioning of the network. Empirical results supporting this hypothesis
are presented in this work. To alleviate this issue, a new attention based FSOD
method is proposed. It computes attention between different feature levels,
enabling a stronger query-support matching. This novel method performs better
than existing methods on small objects and outperforms state-of-the-art on DOTA
and DIOR. 

\noindent
The contributions of this paper are summarized as follows:
\begin{enumerate}
    \item A review of existing works for FSOD with a focus on attention-based
    methods.
    \item A generic and modular structure, named Alignment, Attention, Fusion
    (AAF) framework is designed to implement and compare existing attention
    mechanisms for FSOD.
    \item A comparative performance study based on Relative Mean Average
    Precision (RmAP), a proposed metric for FSOD methods, and an analysis of the
    specific difficulties of applying FSOD on aerial images.
    \item A novel multiscale attention method, named Cross-Scales Query-Support
    Alignment (XQSA) for FSOD, specifically designed for
    small objects. It outperforms previous techniques on aerial images, but also
    on natural images to a lesser extent. 
\end{enumerate}

\section{Review of Existing Work on FSOD}
\label{sec:related}
This section first presents a brief summary of object detection and few-shot
learning. Then, a thorough review of few-shot object detection is conducted,
focusing especially on attention-based methods.

\subsection{Object Detection}
Object detection has made impressive progress with the rise of Convolutional
Neural Networks (CNN). YOLO \cite{redmon2016you} and Faster R-CNN \cite{ren2015faster}
have been the first fully convolutional approaches for object detection. These
two architectures are the most representative examples of the two kinds of
detectors that exist: one-stage and two-stages detectors. Plenty of improvements
have been introduced over these methods. Focal Loss \cite{lin2017focal} improves
training balance between background and foreground objects.
Fully Convolutional One-Stage object detection (FCOS) \cite{tian2019fcos} gets
away from the concept of anchors boxes. Feature Pyramid Network (FPN)
\cite{lin2017feature} improves detection of objects of various sizes.
Recently, various works exploit self-attention mechanisms to increase detection
quality. Dynamic Head \cite{dai2021dynamic} combines three different types of
attention, based on scale, location and current task. Pushing even further, DETR
\cite{carion2020end} replaces the convolutional regressor and classifier with
transformers.

\subsection{Basic Concepts on Few-Shot Learning}
\label{sub:basic}
Few-Shot Learning (FSL) aims at learning tasks from only limited data. The main
principle is to first learn generic knowledge and then adapt the model based on
the limited data available. Generally, the adaptation phase consists in adding
new classes (novel classes) to the problem. We talk about $K$-shots, $N$-ways
learning when only $K$ examples are available for each of the $N$ novel classes.
For a given dataset, it is common to divide the set of classes into base and
novel classes. Base classes are used during training while novel classes are
reserved to assess the generalization capabilities of the models. There exist
different approaches for FSL, we propose a classification of these methods into
four categories: fine-tuning, meta-learning, metric learning and
attention-based. This classification is designed for FSOD and our analysis, but
a more complete taxonomy has already been proposed by
\cite{wang2020generalizing}.


\noindent
\textbf{Fine-tuning \textendash} It consists in training the model on a large
dataset with base classes examples only and then fine-tune it with a few
examples for the novel classes. While conceptually simple, these approaches are
often prone to catastrophic forgetting \cite{kirkpatrick2017overcoming}:
performance drops on base classes after fine-tuning. Fine-tuning on its own is
not very powerful for FSL, but it is part of most other methods as their
training strategy. 

\noindent
\textbf{Meta-learning \textendash} It attempts to learn models that can quickly
adapt to a task. This is often achieved through an \textit{episodic training
strategy}. To mimic FSL testing conditions, at each episode, a subset of classes
$\mathcal{C}_{ep} \subset \mathcal{C}$ is sampled, along with a few examples
divided into support and query sets, containing objects from $\mathcal{C}_{ep}$.
The model is trained on the query set and has access to supplementary
information in the support set. This can be leveraged to choose an
initialization point \cite{finn2017model} or to perform weight updates
\cite{ravi2016optimization}. While the latter methods require the use of an
auxiliary network and therefore do not scale very well, the episodic strategy
helps to train adaptive models and is employed in many other FSL methods.


\noindent
\textbf{Metric Learning \textendash} This has been introduced for FSL by
Prototypical Networks \cite{snell2017prototypical}. It aims at learning an
embedding function from the base dataset, such that the embedding space is
semantically structured. Input images are then classified according to the
distance between their representations and the representations of the support
examples. 

\noindent
\textbf{Attention-based FSL \textendash} It uses support representations to
change the parameters of the model and adapt to new tasks on the fly. Reference
\cite{bertinetto2016learning} proposes a \textit{learnet} whose purpose is to
output weights for the main network based on the support images. Hence, the
network has dynamic convolutional filters and can adapt to new classes. This can
be seen as an attention mechanism between a query image and support images: the
query features are reweighted by the support features through a channel-wise
multiplication. 
Attention is meant to highlight features that are relevant for the task. In the
case of self-attention, this can be achieved by channel multiplication with the
map itself, as proposed by \cite{hu2018squeeze}. It can also be with spatially
distant features of the same image as in non-local neural networks
\cite{wang2018non} or Visual Transformers (ViT) \cite{dosovitskiy2020image}. But
attention can also be computed with features coming from different images. This
can be used for FSL to highlight support features in query features.
For instance, Cross-Transformers \cite{doersch2020crosstransformers}, leverage a
transformer-like attention to align support and query features.

\begin{table*}[h]
    \caption{Comparison of the FSOD methods from an attention perspective. This
    table separates the attention mechanisms into three components: spatial
    alignment, global attention, and fusion layer. All frameworks are working with
    multiscale features except for the one with the mention no FPN. For fusion
    layer, concatenation is denoted as $[\cdot,\cdot]$, while other pointwise
    operations are represented by a circle operator (e.g. $\oplus$ for addition).
    Learnable modules are denoted by $^{\dagger\dagger}$.}
    \label{tab:comparison}
    \begin{adjustbox}{width=\textwidth}
        \rowcolors{2}{gray!25}{white}
        \begin{tabular}{@{}lllllll}
            \toprule[1pt]
            \textbf{Approach}                                                                    & \textbf{Name}                         & \textbf{Date}  & \textbf{Framework}         & \textbf{Alignment}                   & \textbf{Global Attention}    & \textbf{Fusion}                    \\ \hline
            \cellcolor{white}                                                                    & FRW \cite{kang2019few}                & 2019           & YOLO (no FPN)              & None                                  & GP + CRW                    & None                               \\
            \cellcolor{white}                                                                    & RSI   \cite{deng2020few}              & 2019           & YOLO                       & None                                  & GP + CRW                    & None                               \\
            \cellcolor{white}                                                                    & MRCNN \cite{yan2019meta}              & 2019           & Faster R-CNN (no FPN)      & None                                  & GP + CRW                    & None                               \\
            \cellcolor{white}                                                                    & ARMRD \cite{fan2020fsod}              & 2020           & Faster R-CNN               & None                                  & GP + CRW                    & None                               \\
            \cellcolor{white}                                                                    & VEOW  \cite{xiao2020few}              & 2020           & Faster R-CNN               & None                                  & GP + CRW                    & Pool + $[ \odot, \ominus, Id ]$    \\
            \cellcolor{white}                                                                    & WSAAN \cite{xiao2020fsod}             & 2020           & Faster R-CNN               & None                                  & GP + GNN + CRW              & None                               \\
            \cellcolor{white}                                                                    & CACE  \cite{wallach2019one}           & 2020           & Faster R-CNN               & QS Alignment                          & GP + CRW                    & None                               \\
            \cellcolor{white}                                                                    & KT    \cite{kim2020few}               & 2020           & Faster R-CNN               & None                                  & GP + GNN + CRW              & None                               \\
            \cellcolor{white}                                                                    & IFSOD \cite{ganea2021incremental}     & 2021           & Center Ne (no FPN)         & None                                  & GP + GNN + CRW              & None                               \\                                 
            \cellcolor{white}                                                                    & WOFT  \cite{li2020one}                & 2021           & FCOS                       & None                                  & GP + CRW                    & Pool + $[\cdot, \cdot]^{\dagger\dagger}$    \\
            \cellcolor{white}                                                                    & FPDI  \cite{gao2021fast}              & 2021           & Faster R-CNN               & Iterative Alignment via Optimization  & CRW                         & None                               \\
            \cellcolor{white}                                                                    & MFRCN \cite{han2021meta}              & 2021           & Faster R-CNN               & RoI Pooling + QS Alignment            & Global Similarity RW        & $[ \ominus, [\cdot, \cdot]]^{\dagger\dagger}$       \\
            \cellcolor{white}                                                                    & MDETR \cite{zhang2021meta}            & 2021           & DETR (no FPN)              & Transformers-based Alignment          & Transformers Self-Attention & None                               \\
            \cellcolor{white}                                                                    & DRL   \cite{liu2021dynamic}           & 2021           & Faster R-CNN               & None                                  & None                        & Pool + $[\odot, \ominus, \text{Id} ]$   \\
            \cellcolor{white}                                                                    & DANA  \cite{chen2021should}           & 2021           & Faster R-CNN and RetinaNet & QS Alignment                          & Background Attenuation      & $[\cdot, \cdot]$                   \\
            \cellcolor{white}                                                                    & SP    \cite{xu2021few}                & 2021           & Faster R-CNN               & Self-Alignment (query and support)    & None                        & $[\cdot, \cdot]$                   \\
            \cellcolor{white}                                                                    & JCACR \cite{chu2021joint}             & 2021           & YOLO                       & QS Alignment         (higher order)   & None                        & $[\cdot, \cdot]$                   \\ 
            \cellcolor{white}                                                                    & TI \cite{li2021transformation}        & 2021           & Faster R-CNN               & None                                  & None                        & $[\cdot, \cdot]$                   \\
            \multirow{-19}{*}[0mm]{\cellcolor{white}\textbf{Attention}}                          & FCT \cite{han2022few}                 & 2022           & Faster R-CNN               & QS Alignment (with fused keys)        & None                        & $[\cdot, \cdot]$                   \\ \hline
            \cellcolor{white}                                                                    & PNPDet\cite{zhang2021pnpdet}          & 2021           & Center Net (no FPN)        & None                                  & CRW                         & None                               \\
            \cellcolor{white}                                                                    & UPE   \cite{wu2021universal}          & 2021           & Faster R-CNN               & QS Alignment                          & None                        & $\text{Id}  \oplus [\cdot, \cdot]^{\dagger\dagger}$ \\ 
            \multirow{-3}{*}[0mm]{\cellcolor{white}\parbox{1.5cm}{\textbf{Attention/ Metric}}}   & GD \cite{liu2021gendet}               & 2021           & FCOS                       & None                                  & GP + CRW                    & None                               \\ \hline
            \cellcolor{white}                                                                    & RM    \cite{karlinsky2019repmet}      & 2018           & Faster R-CNN               & None                                  & None                        & None                               \\
            \cellcolor{white}                                                                    & RNI \cite{yang2020restoring}          & 2020           & Faster R-CNN               & None                                  & None                        & None                               \\
            \cellcolor{white}                                                                    & FSCE  \cite{sun2021fsce}              & 2021           & Faster R-CNN               & None                                  & None                        & None                               \\
            \cellcolor{white}                                                                    & PFRCN \cite{jeune2021experience}      & 2021           & Faster R-CNN               & None                                  & None                        & None                               \\ 
            \cellcolor{white}                                                                    & AD \cite{cao2021few}                  & 2021           & Faster R-CNN               & None                                  & None                        & None                               \\
            \multirow{-6}{*}{\cellcolor{white}\parbox{1.5cm}{\textbf{Metric Learning}}}          & GDSVD \cite{wu2021generalized}        & 2021           & Faster R-CNN               & None                                  & None                        & None                               \\ \hline
            \cellcolor{white}                                                                    & LSTD  \cite{chen2018lstd}             & 2018           & Faster R-CNN               & None                                  & None                        & None                               \\
            \cellcolor{white}                                                                    & WOFG  \cite{fan2021generalized}       & 2020           & Faster R-CNN               & None                                  & None                        & None                               \\
            \cellcolor{white}                                                                    & TFA   \cite{wang2020frustratingly}    & 2020           & Faster R-CNN               & None                                  & None                        & None                               \\
            \cellcolor{white}                                                                    & MSPSR \cite{wu2020multi}              & 2021           & Faster R-CNN               & None                                  & None                        & None                               \\ 
            \cellcolor{white}                                                                    & DETRG \cite{bar2022detreg}            & 2021           & Faster R-CNN               & None                                  & None                        & None                               \\
            \cellcolor{white}                                                                    & HFSOD \cite{zhang2021hallucination}   & 2021           & Faster R-CNN               & None                                  & None                        & None                               \\
            \cellcolor{white}                                                                    & DHP \cite{wolf2021double}             & 2021           & Faster R-CNN               & None                                  & None                        & None                               \\
            \cellcolor{white}                                                                    & SAM \cite{huang2021few}               & 2021           & Faster R-CNN               & None                                  & Channel + Spatial Attention & None                               \\
            \multirow{-9}{*}{\cellcolor{white}\parbox{1.5cm}{\textbf{Fine-tuning}}}              & SIMPL \cite{xu2022simpl}              & 2021           & Faster R-CNN (no FPN)      & None                                  & None                        & None                               \\ \bottomrule[1pt]
            \end{tabular}%
    \end{adjustbox}

\end{table*}

\subsection{Few-Shot Object Detection}
This section aims at reviewing the FSOD literature. While more emphasis is put
on attention-based methods, it also presents works from other few-shot areas:
fine-tuning, metric learning and meta-learning. Few methods rely solely on
meta-learning, these are included in other subsections. 

\subsubsection{Fine-tuning}
Low Shot Transfer Detector (LSTD) is a pioneer work on FSOD \cite{chen2018lstd}.
It proposes to first train a detector on a base dataset and fine-tune it on a
novel set containing only some examples of the novel classes. Regularization
losses are introduced to prevent catastrophic forgetting. Closely related,
reference \cite{wang2020frustratingly} leverages the same idea without
additional loss. Instead, they freeze most of the network after base training.
Reference \cite{wu2020multi} also proposes a basic fine-tuning strategy by
leveraging a multiscale refinement branch. It provides a better balance between
positive and negative samples and makes both base training and fine-tuning more
efficient. Another proposed method \cite{fan2021generalized} trains two Faster
R-CNN: one on base classes and one on all classes (base and novel), as a
fine-tuned version of the first one. Outputs from both detectors are combined at
test time to achieve better performance on base and novel classes.


\subsubsection{Metric Learning}
Metric-learning based methods are extensively employed for few-shot
classification. For detection however, it remains unusual and therefore neglects
object localization. RepMet \cite{karlinsky2019repmet} learns class
representative vectors to classify Regions of Interest (RoI) in Faster R-CNN.
Closely related, \cite{zhang2021pnpdet} learns prototype vectors as well as
scale factors. This differs slightly from the original metric learning paradigm
as prototypes are learned through training and not computed from examples like
in Prototypical Faster R-CNN \cite{jeune2021experience}. The authors of this
method propose to replace the classification layers in both stage of Faster
R-CNN by prototypical networks. Similarly, FSCE \cite{sun2021fsce} computes the
prototypes directly from the examples but only in the second stage of the
network. In addition, it leverages a contrastive loss function to better
organize the embedding space. Reference \cite{wu2021universal} also uses
prototypes, but as reweighting vectors to enhance class-specific features of the
embedded RoI.

\subsubsection{Attention-based}
\label{sec:attention_fsod}
The main principle of attention-based FSOD is to highlight relevant features for
detection of a particular class, based on examples of that class. A seminal work
in this field is \cite{kang2019few}, which trains a reweighting module along
with a YOLO detector. The reweighting module produces class-specific feature
vectors with a Global Pooling (GP) on the support feature maps. These are then
channel-wise multiplied with the query features extracted by the backbone.
Hence, class-specific query features are generated, and the detection head
computes predictions for each class separately. We denote this operation as
Class Re-Weighting (CRW). This has been used with different detection
frameworks, for instance, references \cite{li2020one} and \cite{xiao2020few} are
built upon Faster R-CNN and FCOS respectively. Other authors proposed to
leverage multiscale reweighting vectors, such as in \cite{deng2020few} for an
application on aerial images. More sophisticated ways to combine information
from query and support have also been proposed. For instance, reference
\cite{fan2020fsod} trains three different heads that link, globally, locally,
and patch-to-patch, the features from the query and the support. Graph Neural
Networks (GNN) can also combine the features and learn semantic relations
between classes (see e.g. \cite{xiao2020fsod, kim2020few}). Another way to
combine query and support is to concatenate the features as in
\cite{dai2021dynamic}.

\begin{figure}[]
    \centering
    \includegraphics[width=\columnwidth, trim=70 0 0 0, clip]{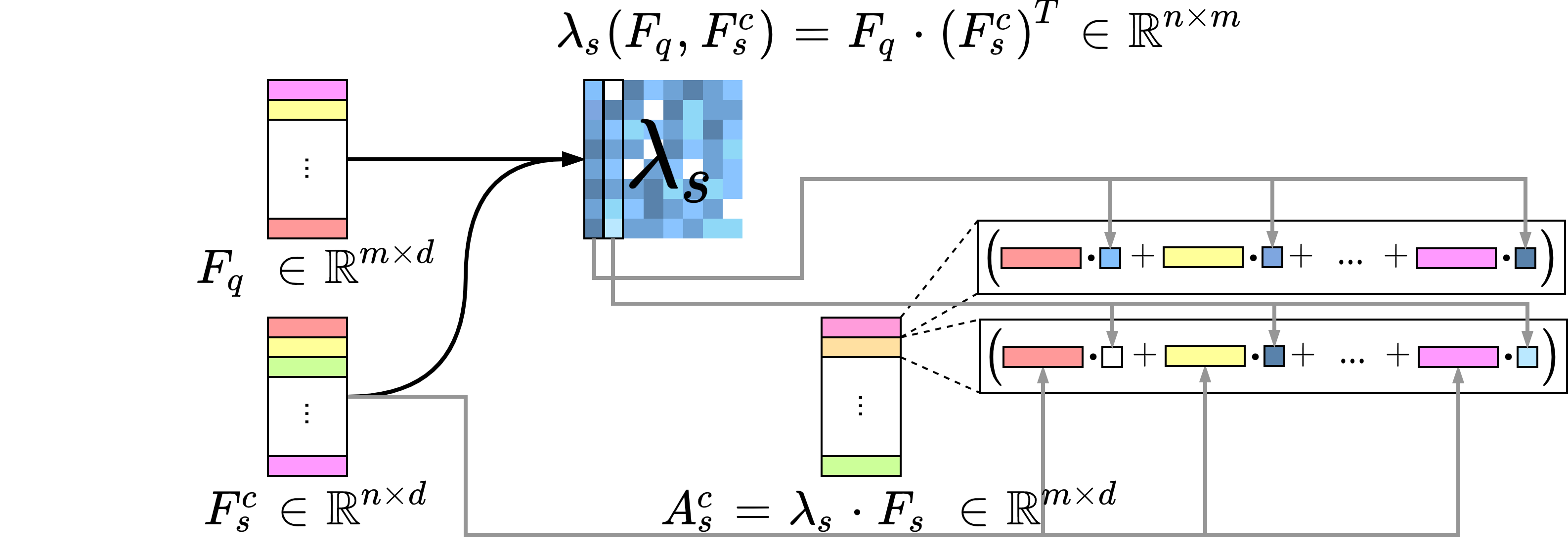}
    \vspace{-5mm}
    \caption{Spatial alignment between query and support feature maps.
     Similarity matrix is based on outer product between the maps. For sake of
     clarity, maps are reshaped as 2-D matrix where the first dimension controls
     the spatial position in the map: $m$ positions for the query and $n$ for
     the support. $d$ is the number of channels. Similar colors mean that
     features are similar.}
     \label{fig:spatial_align}
\end{figure}

The attention-based methods presented above are \textit{global}. The features of
a support example are compressed into a single vector and spatial information is
lost. To prevent this, spatial attention is introduced in several methods. It
can be seen as a spatial realignment of the features. For each location in the
query map, the features from a support example are linearly combined according
to a chosen similarity measure (see \cref{fig:spatial_align}). This highlights
common features between query and support images while preserving spatial
information. We will refer to this as Query-Support Alignment (QSA) in the
following. This kind of alignment has been introduced in \cite{wang2018non} for
video classification. References \cite{chu2021joint, chen2021should} are based
on this approach. Thus, the aligned maps can be easily combined and fed to the
detection head. The combination is often performed through a fusion layer
composed of several point-wise operations. In some cases, additional learnable
modules are included to process the combination of query and support features
before fusion, this is indicated with the symbol $^{\dagger\dagger}$ in
\cref{tab:comparison}. The alignment mechanism can be associated with the global
attention methods mentioned in the previous paragraph as in
\cite{wallach2019one} and \cite{han2021meta}. Alignment can also be carried out
on the feature itself (i.e. without information from the support) as in
\cite{zhang2021meta, xu2021few}. This self-alignment (or self-attention) is at
the heart of transformers. DETR \cite{carion2020end} is a detection framework
based on ViT. It is therefore well suited for this kind of alignment mechanisms.
The few-shot variant of DETR, Meta-DETR described in \cite{zhang2021meta},
combines both self-attention and query-support alignment, achieving impressive
performance.

\cref{tab:comparison} summarizes this literature analysis. Attention mechanisms
are split into the three components that were highlighted in
\cref{sec:attention_fsod}: spatial alignment, global attention and fusion
layer. The table also includes FSOD methods described in previous sections that
do not rely on an attention mechanism.

\section{AAF Framework for Attention in FSOD}
\label{sec:framework}
In \cref{sec:attention_fsod}, three main components of attention mechanisms for
FSOD have been identified: spatial alignment, global attention and fusion layer.
Most attention-based FSOD methods rely on one or more of these components as
shown in \cref{tab:comparison}. To compare them fairly, we propose the
Alignment, Attention and Fusion (AAF) framework. The purpose of this framework
is to provide a flexible environment to implement attention-based FSOD methods.
It takes as input the features from the query image $F_q$ as well as the
features extracted from every support images $F_s^c$ for $c \in \mathcal{C}$
(the set of all classes). It outputs class-specific query features $I_q^c$ in
which features relative to class $c$ are reinforced. To match the three
components of attention described above, the framework is also divided into
three parts as shown in \cref{fig:aaf}. Each component is described below
independently with examples of possible design choices. Even though
this framework is presented from the perspective of object detection, it could
be applied for any kind of few-shot tasks.

\subsection{Query-Support Alignment}
The alignment module, denoted $\Lambda$, spatially aligns the features from the
query and the support. It is unlikely that objects of the same class appear at
the same position inside query and support images. This issue is commonly
avoided by pooling the support map and using it as a class-specific reweighting
vector. But as discussed above, this trick looses the spatial information about
the support object, which can be detrimental for detection. Instead, an
alignment based on attention can be done between query and support feature maps. The
idea is to re-organize one feature map in comparison with the other, so that
similar features are spatially close in the maps. The alignment module is
defined as follows:

\vspace{8mm}
\begin{align}
    \tikzmarknode{g}{\highlight{Purple}{$A_q^c$}} &= \tikzmarknode{l}{\highlight{ForestGreen}{$\lambda_q(F_q , F_s^c)$}} \tikzmarknode{q}{\highlight{red}{$F_q$}},\label{eq:align1}\\
    \tikzmarknode{t}{\highlight{Purple}{$A_s^c$}} &= \tikzmarknode{l}{\highlight{ForestGreen}{$\lambda_s(F_q, F_s^c)$}} \tikzmarknode{s}{\highlight{blue}{$F_s^c$}}.\label{eq:align2}
\end{align}
\begin{tikzpicture}[overlay,remember picture,>=stealth,nodes={align=left,inner ysep=1pt},<-]
    \path (q.north) ++ (0em,1em) node[anchor=south west,color=red!67] (query){$\substack{\text{Query Features } \\ \in \, \mathbb{R}^{n\times d}}$};
    \draw [color=red!57](q.north) |- ([xshift=-0.3ex,color=red]query.north east);
    \path (s.south) ++ (0,-1em) node[anchor=north west,color=blue!67] (support){$\substack{\text{Support Features for class } c \\ \in \, \mathbb{R}^{m\times d}}$};
    \draw [color=blue!57](s.south) |- ([xshift=-0.3ex,color=blue]support.south east);
    \path (l.south) ++ (0,-1em) node[anchor=north east,color=ForestGreen!80] (lambda){$\substack{\text{Affinity matrices } \\ \in \, \mathbb{R}^{n\times m}}$};
    \draw [color=ForestGreen!80](l.south) |- ([xshift=-0.3ex,color=ForestGreen]lambda.south west);
    \path (g.north) ++ (0,1em) node[anchor=south east,color=Purple!80] (gfeat){$\substack{\text{Aligned features }}$};
    \draw [color=Purple!80](g.north) |- ([xshift=-0.3ex,color=Purple]gfeat.north west);
\end{tikzpicture}
\vspace{2\baselineskip}

The definition of the matrices $\lambda_q$ and $\lambda_s$ determines how
features are aligned. This formulation is quite similar to the non-local blocks
described in \cite{wang2018non} and is at the heart of visual transformers
\cite{vaswani2017attention}. Transformers attention can be understood as an
alignment of the value to match the query-key similarity. This formulation
allows implementing easily different feature alignments by changing the
expression of the affinity matrices. As an example, Meta Faster R-CNN
\cite{han2021meta} leverages an alignment module with affinity matrices
$\lambda_s(F_q , F_s^c) = F_q\cdot (F_s^q)^T$ and $\lambda_q(F_q , F_s^c) = I$
(see Example A in \cref{fig:aaf}). Only the support features are aligned so that
they match query features. 

\subsection{Global Attention}
The global attention module, denoted  $\Gamma$, combines global information of the supports
and the query. It highlights class-specific features and softens irrelevant
information for the task. This module is defined as follows:

\vspace{2mm}
\begin{align}
    \tikzmarknode{g}{\highlight{darkgray}{$H_q^c$}} &= \tikzmarknode{l}{\highlight{RedOrange}{$\gamma_q$}}(A_q^c , A_s^c),\label{eq:global1}\\
    \tikzmarknode{g}{\highlight{darkgray}{$H_s^c$}} &= \tikzmarknode{l}{\highlight{RedOrange}{$\gamma_s$}}(A_q^c, A_s^c).\label{eq:global2}
\end{align}
\begin{tikzpicture}[overlay,remember picture,>=stealth,nodes={align=left,inner ysep=1pt},<-]
    \path (l.south) ++ (0,-1em) node[anchor=north west,color=RedOrange!80] (lambda){\footnotesize Global attention operators};
    \draw [color=RedOrange!80](l.south) |- ([xshift=-0.3ex,color=RedOrange]lambda.south east);
    \path (g.south) ++ (0,-1em) node[anchor=north east,color=darkgray!80] (gfeat){\footnotesize Highlighted features};
    \draw [color=darkgray!80](g.south) |- ([xshift=-0.3ex,color=darkgray]gfeat.south west);
\end{tikzpicture}
\vspace{2\baselineskip}

The global attention operators $\gamma_q$, $\gamma_s$ combine the global
information from their inputs and highlight features accordingly. This is
generally done through channel-wise multiplication. In this way, class-specific
features are highlighted, while features not relevant to the class are softened.
Changing the definition of $\gamma_q$ and $\gamma_s$ allows the implementation
of a wide variety of global attention mechanisms. For instance, reference
\cite{kang2019few} pools the support maps with a global max pooling operation
($\text{GP}$) into a reweighting vector and reweights the query features
channels with it: $\gamma_q(A_q^c, A_s^c) = A_q^c \circledast GP(A_s^c)$  and
$\gamma_s(A_q^c, A_s^c) = A_s^c$ (see Example B in \cref{fig:aaf}).

\subsection{Fusion Layer}
The purpose of the fusion component is to combine highlighted query and support
maps. This is only applicable when the maps have the same spatial dimension. It
is mostly used alongside with the alignment module as the latter does not combine the
information from the support and the query but only reorganizes the maps. In
particular, when support and query maps do not have the same spatial dimension,
aligning support maps with query maps fixes the size discrepancy. The fusion
module is defined as follows:

\begin{equation}
    \label{eq:fusion}
    \tikzmarknode{g}{\highlight{TealBlue}{$M_q^c$}} = \tikzmarknode{l}{\highlight{Mahogany}{$\Omega$}}(H_q^c , H_s^c).
\end{equation}
\begin{tikzpicture}[overlay,remember picture,>=stealth,nodes={align=left,inner ysep=1pt},<-]
    \path (l.south) ++ (0,-1em) node[anchor=north west,color=Mahogany!80] (lambda){\footnotesize Fusion operator};
    \draw [color=Mahogany!80](l.south) |- ([xshift=-0.3ex,color=Mahogany]lambda.south east);
    \path (g.south) ++ (0,-0.9em) node[anchor=north east,color=TealBlue!80] (gfeat){\footnotesize Merged query features};
    \draw [color=TealBlue!80](g.south) |- ([xshift=-0.3ex,color=TealBlue]gfeat.south west);
\end{tikzpicture}
\vspace{1\baselineskip}

\begin{figure*}[h]
    \centering
    \includegraphics[width=0.9\textwidth]{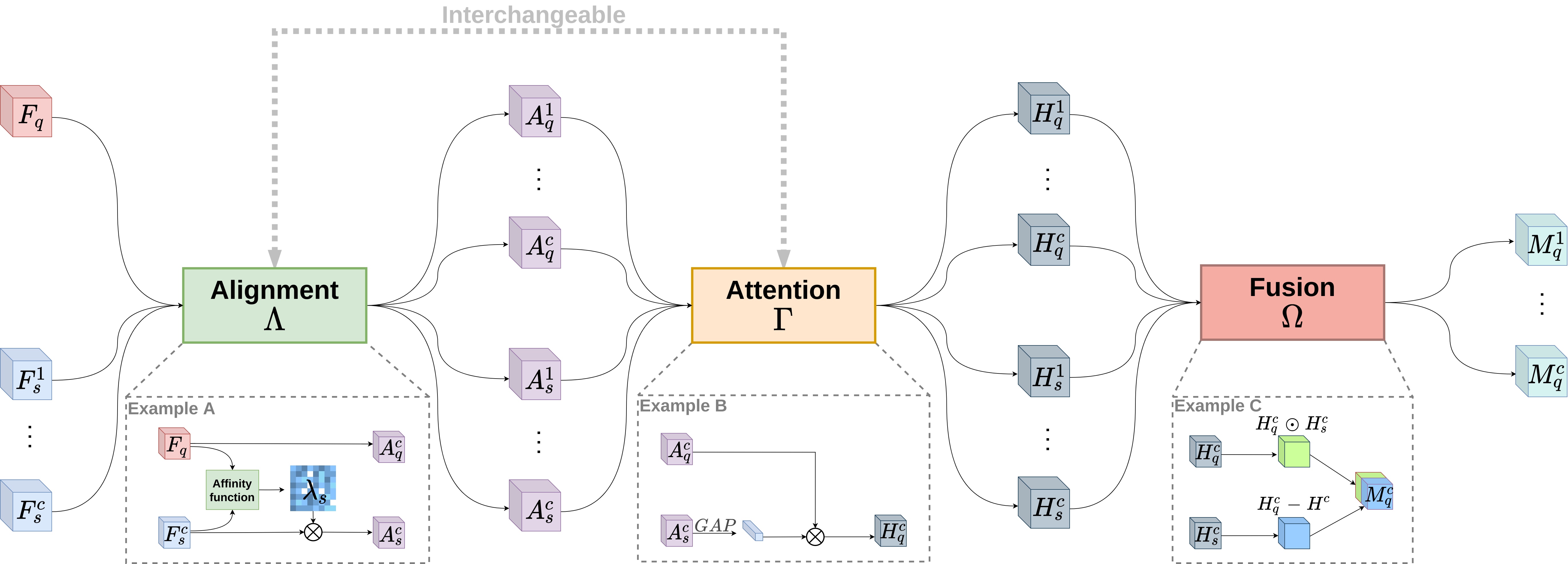}
    \caption{The Alignment Attention Fusion (AAF) framework is composed of three
    components: spatial alignment $\Lambda$, global attention $\Gamma$ and a
    fusion layer $\Omega$. Examples for each module are depicted, these come
    from FSOD methods in the literature. Example A is presented in
    \cite{han2021meta}, Example B in \cite{kang2019few} and Example C in
    \cite{liu2021dynamic}. The colors chosen in this diagram match the colors
    used in \Cref{eq:align1,eq:align2,eq:global1,eq:global2,eq:fusion}.}
    \label{fig:aaf}
\end{figure*}

The highlighted maps can be combined through any point-wise operation, addition
$\oplus$, multiplication $\odot$, subtraction $\ominus$, concatenation $[\cdot,
\cdot ]$ or more sophisticated ones. An example of such a fusion module is
presented in \cite{liu2021dynamic}. The fusion operator concatenates the results
of the addition and the subtraction of the highlighted features: $\Omega(H_q^c,
H_s^c) = [H_q^c \oplus H_s^c, H_q^c \ominus H_s^c]$ (see Example C in
\cref{fig:aaf}). The point-wise operators can also contain small trainable
models such as in \cite{han2021meta}, where small CNNs are applied after the
point-wise operators, but before the concatenation: $\Omega(H_q^c, H_s^c) =
[\psi_{dot}(H_q^c \odot H_s^c), \psi_{sub}(H_q^c \ominus H_s^c),
\psi_{cat}([H_q^c, H_s^c])]$.

Except for the fusion layer which must be applied last, spatial alignment and
global attention can be applied in any order. This flexibility is required to
implement methods that apply global attention before alignment, such as DANA
\cite{chen2021should}. The author proposed a so-called background attenuation
block that behaves like a self-attention block to refine support features. They
applied it on support features before alignment. However, similar technique
could very well be applied to query feature after alignment with support. Hence,
the interchangeability is required to offer more flexibility to the framework.
The whole architecture of the AAF framework is illustrated in \cref{fig:aaf}, in
which examples from the previous sections are also depicted. The modular
structure of the framework enables the implementation of a wide variety of
attention mechanisms while keeping all other hyperparameters fixed. It is
therefore a useful tool for FSOD methods comparison.

\subsection{Implementation details}
In order to make the comparison fair, some implementation details are kept fixed
for all experiments. The backbone is a ResNet-50 with a 3-layers Feature Pyramid
Network on top. It extracts features at 3 different levels, which helps the
network to detect objects of various sizes. The detection head is based on FCOS
\cite{tian2019fcos}. Each feature level is processed by the same head to output
detections of different sizes. The AAF framework is applied between the backbone
and the detection head. As features are extracted at multiple levels, attention
mechanisms are also implemented to work at different scales. This may differ
from the original implementations, but most methods are already designed to work
at multiscale (see \cref{tab:comparison}). The networks are trained in an
episodic manner. During each episode, a subset $\mathcal{C}_{ep} \subset
\mathcal{C}$ of the classes is randomly sampled ($\lvert\mathcal{C}_{ep} \rvert=
3$). Only the annotations of the episode classes are used for training the
model. A support set is sampled at each episode containing $K$ examples for each
episode class. A query set is also sampled for each episode, containing 100
images per class, hence 300 images are given to the model at each episode.

The training is divided in two phases \textit{base training} and
\textit{fine-tuning}. During base training, only base classes can be sampled
($\mathcal{C}_{ep} \subset \mathcal{C}_{base}$) and one example per class is
drawn for the support set ($K=1$). The optimization is done with SGD and a
learning rate of $\num{1e-3}$ for 1000 episodes. During \textit{fine-tuning},
the backbone is frozen, the learning rate is divided by 10 and the episode
classes can be sampled from $\mathcal{C}_{base} \cup \mathcal{C}_{novel}$, with
at least one novel class per episode. Examples from novel classes are selected
among the $K$ examples sampled once before fine-tuning. Each model is fine-tuned
separately for different value of $K \in {1,3,5,10}$. The base/novel class
splits are selected randomly before training. 3 classes are selected as novel
classes for DOTA, 5 for Pascal VOC and DIOR and 20 for MS COCO. The lists of the
selected classes in our experiments are detailed in Appendix
\ref{app:class_split}. During both training phases, the same loss function is
optimized, as defined in FCOS.

The evaluation is also conducted in an episodic manner, following
recommendations from \cite{huang2021survey}. The performance is averaged over
multiple episodes each containing 500 examples for each class and this is
repeated multiple times with randomly sampled support sets. The query and
support examples are drawn from test set, thus the support examples are
different from the ones used during fine-tuning. This prevents overestimations
of the performance due to overfitting on the support examples. 

Some existing works leverage sophisticated training strategies (e.g. auxiliary
loss functions \cite{fan2020few}, hard examples mining
\cite{zhang2021hallucination} or multiscale training \cite{deng2020few}). While
this certainly improves the quality of the detections, it introduces new
parameters to tune and makes the comparison with other works difficult. As the
focus of this study is about attention mechanisms, we chose not to reimplement
all these improvements. However, to remain competitive with existing works, we
propose a novel augmentation pipeline specifically designed for object
detection. Its definition is described in Appendix \ref{app:augmentations},
which includes a cumulative study of its different components on DOTA. In
addition, we chose the support extraction strategy among a few possibilities
detailed in Appendix \ref{app:cropping}. From our analysis, it seems that this
design choice influences significantly the model performance (see
\cref{tab:cropping_methods}). However, it is barely discussed in the FSOD
literature. We find that the best strategy is to crop the support example and
resize it to a fixed size patch. This strategy is thus fixed for the all our
experiments.

\section{Attention-based FSOD Comparison}
\label{sec:comparison}
To showcase the flexibility of the proposed AAF framework, we compare multiple
existing works. Some methods described in \cref{sec:related} are selected: FRW
\cite{kang2019few}, WSAAN \cite{xiao2020fsod}, DANA \cite{chen2021should}, MFRCN
\cite{han2021meta} and DRL \cite{liu2021dynamic} (see \cref{tab:comparison}).
These have been chosen because they represent well the variety of attention
mechanisms available in the literature. FRW is based on class-specific
reweighting vectors, WSAAN has a more sophisticated global attention and
computes reweighting vectors inside a graph structure. DANA and MFRCN leverage
query-support alignment in different manners and DRL only uses a sophisticated
fusion layer. Each of these methods has been reimplemented within the AAF
framework. Of course, some details differ from the original implementations, but
the purpose of this comparison is to compare only the attention mechanisms. In
particular, the backbone and the training strategy (losses and episode tasks)
may differ. We first conduct such a comparative experiment on Pascal VOC
\cite{everingham2010pascal} and MS COCO\cite{lin2014microsoft} datasets. On
these datasets, the performance achieved by our implementations is close to the
values reported in the original papers (see Appendix \ref{app:res_orig}). Then,
we use the framework to compare the performance of some methods on two aerial
datasets:  DOTA \cite{xia2018dota} and DIOR \cite{li2020object}. Finally, the
following sections analyze the FSOD performance discrepancies on aerial and
natural images.


\subsection{Performance Analysis on Natural Images}
\label{sec:res_voc}

First, compared to FCOS baseline on natural images, a slight performance drop on
base classes is visible in \cref{tab:result_voc} (FCOS achieves $0.66$
$\text{mAP}_{0.5}$ on Pascal VOC, all FCOS results are included in
\cref*{tab:rmap_values} in Appendix \ref{app:rmap}). This is expected, even if
the model has seen a lot of examples of these classes during training, its
predictions are still conditioned on a few examples, which can sometimes be
misleading. On the other hand, performance on novel classes is significantly
lower than the FCOS baseline, especially for low numbers of shots. The number of
shots is crucial for performance on novel classes. The higher the number of
shots, the better the network performs. In average, with 10 examples per class,
the network achieves $0.2$ higher mAP than with 1 example. More examples provide
more precise and robust class representations, improving the detection. The same
phenomenon is observed with base classes to a lesser extent ($+ 0.04$ mAP from 1
to 10 shots). \cref{fig:method_compare} in Appendix \ref{app:map_shot_class}
displays these trends clearly, both for base and novel classes. In addition,
\cref{fig:map_per_class} in the same appendix, provides the same results split
by class.

The behavior just described is expected from any few-shot object detection
method. Moreover, performance values are close to what is reported in the
original papers of the reimplemented methods (see
\cref{tab:res_orig_pascal,tab:res_orig_coco} in Appendix \ref{app:res_orig}).
These are not the exact same values as many architectural choices differ from
the proposed methods (e.g. backbone, classes splits, etc.). Nevertheless, it
confirms that the proposed AAF framework is flexible enough to implement a wide
variety of attention mechanisms. It is therefore an appropriate tool to compare
and design query-support attention mechanisms.
\begin{table*}[t]
    \centering
    \caption{Performance comparison between five selected methods (see
    \cref{sec:comparison}) on Pascal VOC. All are reimplemented with the
    proposed AAF framework. Mean average precision is reported for each method
    on base and novel classes separately and for various numbers of shots ($K
    \in \{1,3,5,10\}$). For comparison, the FCOS baseline achieve 0.66
    $\text{mAP}_{0.5}$ on Pascal VOC.}
    \label{tab:result_voc}
    \begin{tabular}{@{}cccccccccccccccc@{}}
    \toprule[1pt]
                      &  & \multicolumn{2}{c}{\textbf{FRW} \cite{kang2019few}} &  & \multicolumn{2}{c}{\textbf{WSAAN} \cite{xiao2020fsod}} &  & \multicolumn{2}{c}{\textbf{DANA} \cite{chen2021should}} &  & \multicolumn{2}{c}{\textbf{MFRCN} \cite{han2021meta}} &  & \multicolumn{2}{c}{\textbf{DRL} \cite{liu2021dynamic}} \\ \hline
    $\boldsymbol{K}$  &  & \underline{Base}         & \underline{Novel}        &  & \underline{Base}        & \underline{Novel}            &  & \underline{Base}      & \underline{Novel}               &  & \underline{Base}         & \underline{Novel}          &  & \underline{Base}    & \underline{Novel}                \\
    1                 &  & 0.599                    & 0.282                    &  & 0.617                   & 0.309                        &  & 0.626                 & \textbf{0.328}                  &  & 0.578                    & 0.302                      &  & \textbf{0.642}      & 0.270                            \\
    3                 &  & 0.633                    & 0.311                    &  & 0.635                   & \textbf{0.422}               &  & \textbf{0.642}        & 0.340                           &  & 0.587                    & 0.368                      &  & 0.617               & 0.296                            \\
    5                 &  & 0.643                    & 0.463                    &  & 0.647                   & \textbf{0.462}               &  & 0.652                 & 0.426                           &  & 0.621                    & 0.408                      &  & \textbf{0.664}      & 0.373                            \\
    10                &  & 0.632                    & 0.487                    &  & 0.653                   & \textbf{0.517}               &  & 0.650                 & 0.503                           &  & 0.634                    & 0.494                      &  & \textbf{0.670}      & 0.480                            \\ \bottomrule[1pt]
    \end{tabular}%
    \end{table*}


DRL is arguably the simplest method among the five selected as it leverages only
a fusion layer. It combines query features with the features of each support
image through concatenation and point-wise operations, creating class-specific
query features. It is therefore the closest to the regular FCOS functioning.
This explains the very good performance on base classes and lower mAP on novel
classes, compared to the baseline. Regarding the other methods, FRW and WSAAN
can be easily compared as both are based on global attention. The only
difference is how the class-specific vectors are computed. In FRW, they are
globally pooled from the support feature map. However, WSAAN combines the same
vectors with query features in a graph. This certainly provides better
class-specific features and in the end, better results both on base and novel
sets. The remaining methods, DANA and MFRCN both leverage spatial alignment.
While it seems to bring quite an improvement for DANA over FRW and DRL, the gain
is smaller for MFRCN. In both methods, spatial alignment is not used alone. It
is combined with other attention mechanisms. In DANA, a Background Attenuation
block (i.e. a global self-attention mechanism) is applied to the support features to
highlight class-relevant features and soften background ones. In MFRCN, aligned
features are reweighted with global vectors computed from the similarity matrix
between query and support features. This last operation may be redundant as the
similarity information is already embedded into the aligned features, whereas
background attenuation leverages new information. 

From this comparison, one can conclude that both global attention and spatial
alignment are beneficial for FSOD. However, these improvements may not always be
compatible, as shown by the results of MFRCN. Hence, the design of each
component must be done carefully so that spatial alignment, global attention,
and fusion work in unison.

\begin{table}[b]
    \centering
    \caption{Performance comparison between WSAAN \cite{xiao2020fsod} and DANA
    \cite{chen2021should} on MS COCO. $\text{mAP}_{0.5:0.95}$ (MS COCO mAP, with
    IoU thresholds ranging from 0.5 to 0.95) and $\text{mAP}_{0.5}$ values are
    reported for base and novel classes separately and for different numbers of
    shots: $K \in \{1, 5, 10, 30\}$.}
    \label{tab:result_coco}
    \resizebox{\columnwidth}{!}{%
    \begin{tabular}{@{}ccccccccccccc@{}}
    \toprule[1pt]
    &                        &  & \multicolumn{4}{c}{\textbf{WSAAN} \cite{xiao2020fsod}}                                   &  & \multicolumn{4}{c}{\textbf{DANA} \cite{chen2021should}}                                     \\
                             &  & \multicolumn{2}{c}{$\text{mAP}_{0.5}$} &  &  \multicolumn{2}{c}{$\text{mAP}_{0.5:0.95}$} &  & \multicolumn{2}{c}{$\text{mAP}_{0.5}$}  &  & \multicolumn{2}{c}{$\text{mAP}_{0.5:0.95}$}    \\ \midrule
    $\boldsymbol{K}$         &  & \underline{Base}  &  \underline{Novel} &  & \underline{Base}        & \underline{Novel}  &  & \underline{Base}   & \underline{Novel}  &  & \underline{Base}   & \underline{Novel}         \\
    1                        &  & 0.335             & 0.120              &  & 0.201                   & 0.066              &  & \textbf{0.355}     & \textbf{0.145}     &  & \textbf{0.213}     & \textbf{0.078}            \\
    5                        &  & 0.399             & 0.199              &  & 0.236                   & 0.105              &  & \textbf{0.428}     & \textbf{0.222}     &  & \textbf{0.252}     & \textbf{0.119}            \\
    10                       &  & 0.409             & 0.214              &  & 0.244                   & 0.115              &  & \textbf{0.430}     & \textbf{0.237}     &  & \textbf{0.256}     & \textbf{0.129}            \\
    30                       &  & 0.415             & 0.222              &  & 0.247                   & 0.121              &  & \textbf{0.435}     & \textbf{0.244}     &  & \textbf{0.260}     & \textbf{0.133}            \\ \bottomrule[1pt]
    \end{tabular}%
    } 
    \end{table}

\begin{table*}[h]
    \centering
    \caption{Comparison of $\text{mAP}_{0.5}$ of several methods on DOTA and
    DIOR datasets. For each method, mAP is reported for different number of
    shots $K \in \{1, 3, 5, 10\}$ and separately for base and novel classes.
    Blue and red values represent the best performance on base and novel classes
    respectively, for each dataset. Methods marked with a
    $^\dagger$ are not re-implemented in the AAF framework, their results are taken
    from the original papers. Our results in this table were published in a
    preliminary conference article \cite{lejeune2022improving}, accepted at EUSIPCO 2022.}
    \label{tab:result_aerial}
    \resizebox{\textwidth}{!}{%
    \begin{tabular}{@{}ccccccccccccccccccccccccccc@{}}
        \toprule[1pt]
        \multicolumn{1}{l}{} & \multicolumn{11}{c}{\textbf{DOTA}}                                                                                                                                                                                              &                      & \multicolumn{14}{c}{\textbf{DIOR}}                                                                                                                                                                                                                     \\ \cmidrule(lr){2-12} \cmidrule(lr){14-27}
                             & \multicolumn{2}{c}{\textbf{FRW}}                      & \textbf{} & \multicolumn{2}{c}{\textbf{WSAAN}}                                        & \textbf{} & \multicolumn{2}{c}{\textbf{DANA}}                                       & \textbf{} & \multicolumn{2}{c}{\textbf{PFRCN$^\dagger$}}   &                      & \multicolumn{2}{c}{\textbf{FRW}}                     & \textbf{} & \multicolumn{2}{c}{\textbf{WSAAN}}                     & \textbf{} & \multicolumn{2}{c}{\textbf{DANA}}                                       & \textbf{} & \multicolumn{2}{c}{\textbf{PFRCN$^\dagger$}}  &  & \multicolumn{2}{c}{\textbf{WSAAN$^\dagger$}}                       \\ \midrule
        $\boldsymbol{K}$     & \underline{Base} & \underline{Novel}                  &           & \underline{Base}                     & \underline{Novel}                  &           & \underline{Base}                   & \underline{Novel}                  &           & \underline{Base} & \underline{Novel}                            &                      & \underline{Base} & \underline{Novel}                 &           & \underline{Base}                    & \underline{Novel}&           & \underline{Base}                   & \underline{Novel}                  &           & \underline{Base} & \underline{Novel}                           &  & \underline{Base}   & \underline{Novel}                                     \\
        1                    & 0.47             & \textcolor{red}{\textbf{0.13}}     &           & 0.46                                 & 0.12                               &           & \textcolor{blue}{\textbf{0.49}}    & \textcolor{red}{\textbf{0.13}}     & \textbf{} & 0.29             & 0.08                                         &                      & 0.57             & 0.17                              &           & 0.56                                & 0.16             &           & \textcolor{blue}{\textbf{0.59}}    & \textcolor{red}{\textbf{0.21}}     &           & 0.41             & 0.06                                        &  & -                  & -                                                     \\
        3                    & 0.47             & \textcolor{red}{\textbf{0.25}}     &           & 0.44                                 & 0.24                               &           & \textcolor{blue}{\textbf{0.50}}    & 0.21                               &           & 0.34             & 0.10                                         &                      & 0.58             & 0.25                              &           & 0.52                                & 0.14             &           & \textcolor{blue}{\textbf{0.59}}    & \textcolor{red}{\textbf{0.27}}     &           & 0.40             & 0.08                                        &  & -                  & -                                                     \\
        5                    & 0.49             & 0.30                               &           & 0.48                                 & \textcolor{red}{\textbf{0.31}}     &           & \textcolor{blue}{\textbf{0.53}}    & 0.25                               &           & 0.32             & 0.09                                         &                      & 0.61             & 0.33                              &           & 0.61                                & 0.30             &           & \textcolor{blue}{\textbf{0.62}}    & \textcolor{red}{\textbf{0.34}}     &           & 0.42             & 0.09                                        &  & -                  & 0.25                                                  \\
        10                   & 0.49             & \textcolor{red}{\textbf{0.37}}     &           & 0.47                                 & 0.35                               &           & \textcolor{blue}{\textbf{0.53}}    & 0.34                               &           & 0.35             & 0.11                                         &                      & 0.62             & \textcolor{red}{\textbf{0.36}}    &           & \textcolor{blue}{\textbf{0.62}}     & 0.33             &           & \textcolor{blue}{\textbf{0.62}}    & \textcolor{red}{\textbf{0.36}}     &           & 0.42             & 0.09                                        &  & 0.54               & 0.32                                                  \\ \bottomrule[1pt]
        \end{tabular}%
    }
    \end{table*}

Another set of experiments is conducted on MS COCO dataset. Only the two
best-performing methods on Pascal VOC are selected and trained on MS COCO
following the same experimental setup. The results are summarized in
\cref{tab:result_coco}. The mAP values are reported following standards from
Pascal VOC ($\text{mAP}_{0.5}$ with one IoU threshold), and MS COCO
($\text{mAP}_{0.5:0.95}$ with several thresholds). MS COCO is a much more
difficult detection benchmark, therefore the numbers of shots are adjusted: 1, 5,
10, and 30 shots. These results comfort the conclusion obtained on Pascal VOC:
the framework is flexible enough to implement various FSOD techniques that
achieve competitive results with state-of-the-art. As for Pascal VOC the models
achieve better performance with more shots. Unlike Pascal VOC, base classes
also benefit significantly from a larger number of examples on MS COCO. MS COCO
being more difficult, the information extracted from the supports better helps
the models. Finally, WSAAN outperforms DANA on Pascal VOC but performs slightly
worse on MS COCO. It can be noted that, the results on a dataset cannot be
extrapolated to another one. A method that performs best on a dataset is not
guaranteed to do so on another dataset. This reinforces the need of a flexible
framework that allows fair and easy comparison between FSOD methods. That way,
the best performing method can be easily selected for a given problem.

From these experiments on natural images, it seems clear that DANA is the best
performing method among the ones tested. Therefore, it highlights the importance
of feature alignment for query-support matching. Global attention looses spatial
information in support features which is detrimental for detection. However,
global attention methods should not be overlooked. WSAAN shows impressive
performance and even outperforms slightly DANA on Pascal VOC. It could be
interesting to combine both methods, but this does not seem to be trivial as
demonstrated by the results of MFRCN which basically leverages the alignment from
DANA and the attention from FRW, but does not yield satisfactory results.

\subsection{Performance Analysis on Aerial Images and comparison with natural images}
\label{sec:res_aerial}

To our knowledge, only three existing works evaluate FSOD methods on aerial
images: FRW \cite{xiao2020few}, WSAAN \cite{deng2020few} and PFRCN
\cite{jeune2021experience}. All these methods are evaluated on different
datasets, making their performance comparison difficult. Using the proposed AAF
framework, we compare the performance of these methods both on DOTA and DIOR.
These methods are reimplemented inside the framework and all common parameters
are fixed during the experiments. In addition, DANA \cite{chen2021should} is
also included in the comparison as it was the best performing method on MS COCO.
\cref{tab:result_aerial} regroups the results of these experiments. These
results show a slight improvement over the state-of-the-art on DIOR (WSAAN
\cite{deng2020few}). Our implementation of WSAAN outperforms (0.08 mAP on base
classes and 0.01 on novel classes) the result reported in the original paper.
However, the attention mechanism employed in WSAAN is not optimal for aerial
images. WSAAN is outperformed by both FRW and DANA. While this was expected for
DANA in the light of results from \cref{sec:res_voc}, it was not for FRW. The
superiority of DANA over the other methods on DOTA and DIOR is clear and
coherent with the results on natural images. The more sophisticated attention
mechanism, in particular the alignment, from DANA is better at extracting and
leveraging the information from the support examples. Hence, the detection
performance is higher. It is particularly beneficial for small number of shots:
the extracted information is semantically robust. 

\begin{figure}[t]
    \centering
    \includegraphics[width=0.95\columnwidth, trim=0 30 0 0, clip]{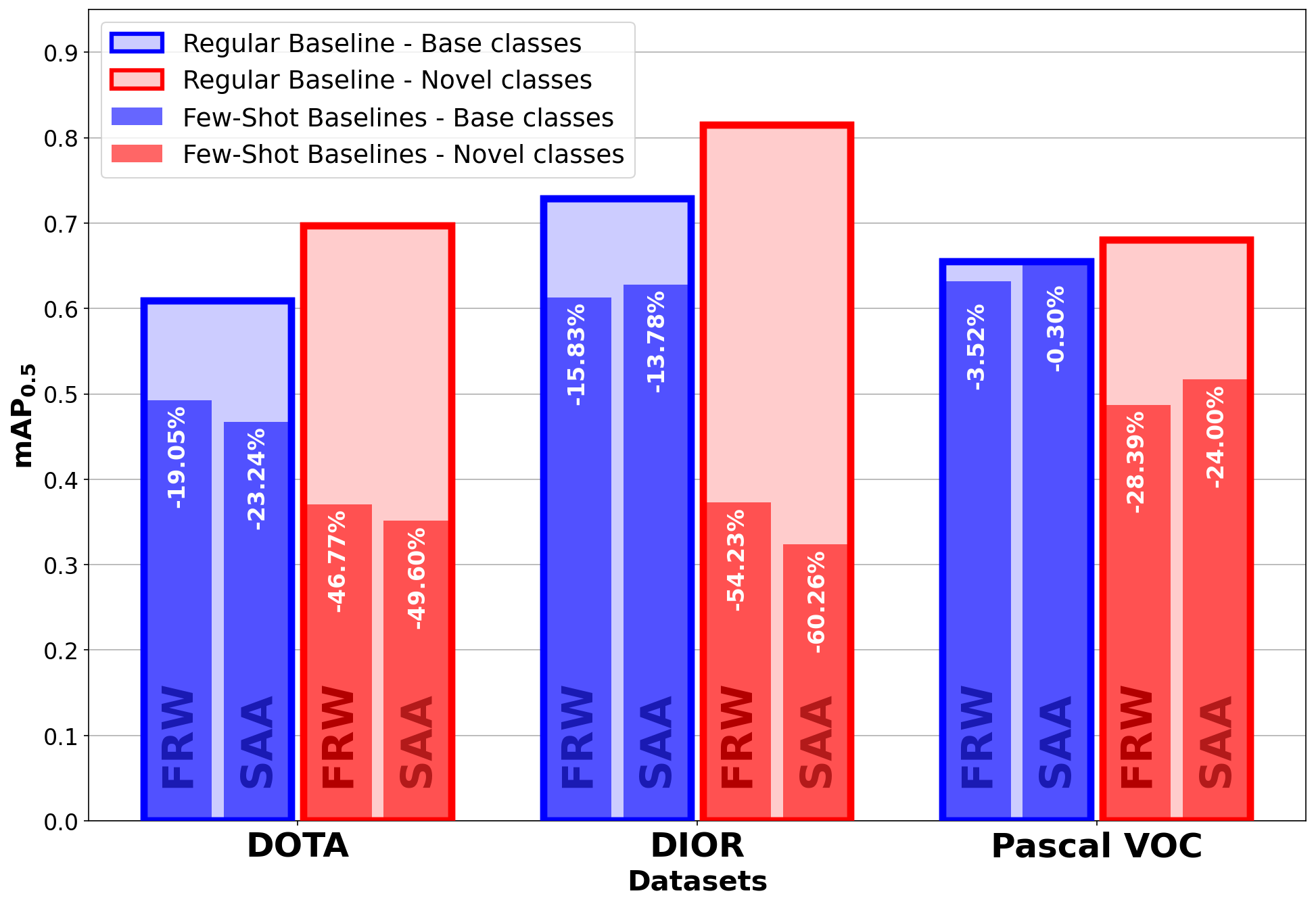}
    \caption{Performance comparison between Regular Baseline and Few-Shot
    Baselines, FRW \cite{deng2020few} and WSAAN (WSA) \cite{xiao2020fsod} on three
    datasets: DOTA, DIOR and Pascal VOC. On aerial image datasets, a large
    performance gap is observed on novel classes, while this is relatively
    reduced on Pascal VOC (i.e.natural images).}
    \label{fig:baseline_res}
    \vspace{-3mm}
\end{figure}

However, it is surprising to see FRW performing so well, even better than DANA on
DOTA's novel classes. On Pascal VOC, which is similar to DOTA and DIOR in terms
of size and number of classes, FRW performs largely worse than DANA. 
In addition to this inconsistency, another result is unsettling: the
performance gap between the classical baseline (i.e. FCOS) and the few-shot
approaches. On natural images, the performance drop is nearly inexistent for
base classes and around 25\% for novel classes. On aerial images these drops are
largely increased: $\sim 15\%$ and $\sim 50\%$ for base and novel classes
respectively. This can be guessed from \cref{tab:result_voc,tab:result_aerial},
but detailed gaps are provided in \cref{fig:baseline_res}. Note that the FSOD
baselines were trained here without augmentation, which explains slightly lower
values than in tables from \cref{sec:res_voc,sec:res_aerial}.

It seems tempting here to extrapolate the FSOD performance on DIOR from the
performance on Pascal VOC. The classical baseline (FCOS) achieves similar
performance on these two datasets, which contain the same number of classes and
roughly the same number of images. Therefore, one could have expected close FSOD
performance on these datasets. This is quite different from the actual results
reported in \cref{fig:baseline_res}. It is generally irrelevant to compare the
performance of a method from a dataset to another, especially with images of
different nature. Each dataset have its own characteristics (resolution,
intra-class variety, color range, etc.) and therefore a given model will not
perform equally good on two distinct datasets with respect to a pre-defined
performance metric. Hence, we cannot compare the absolute performance of a FSOD
method on Pascal VOC and DOTA and the latter extrapolation is not valid.  

Nevertheless, there is a pattern: FSOD methods work consistently better on
natural images compared to aerial images. To understand this phenomenon, we need a
way to fairly compare the FSOD performance across several datasets. To this end,
we propose to look at the relative performance of the FSOD methods against the
non few-shot baseline (i.e. FCOS in our case) using the following metric: 

\begin{equation}
    \label{eq:rmap}
    \text{RmAP} =
            \frac{\text{mAP}_{\text{FSOD}} -\text{mAP}_{\text{Baseline}}}{\text{mAP}_{\text{Baseline}}}.
\end{equation}

This metric assesses how well a FSOD method is performing on different
datasets even if the classical detection performance differs. Hence, it
represents how much performance is lost when switching from regular to the
few-shot regime. This is exactly what is illustrated in \cref{fig:baseline_res},
white percentages are $\text{RmAP}$ values. $\text{RmAP}$ is significantly lower
on DOTA and DIOR compared to Pascal VOC. This way, we can quantitatively confirm the
intuition emerging from the above results: FSOD works better on natural images. 

\begin{figure}
    \centering
    \includegraphics[width=\columnwidth]{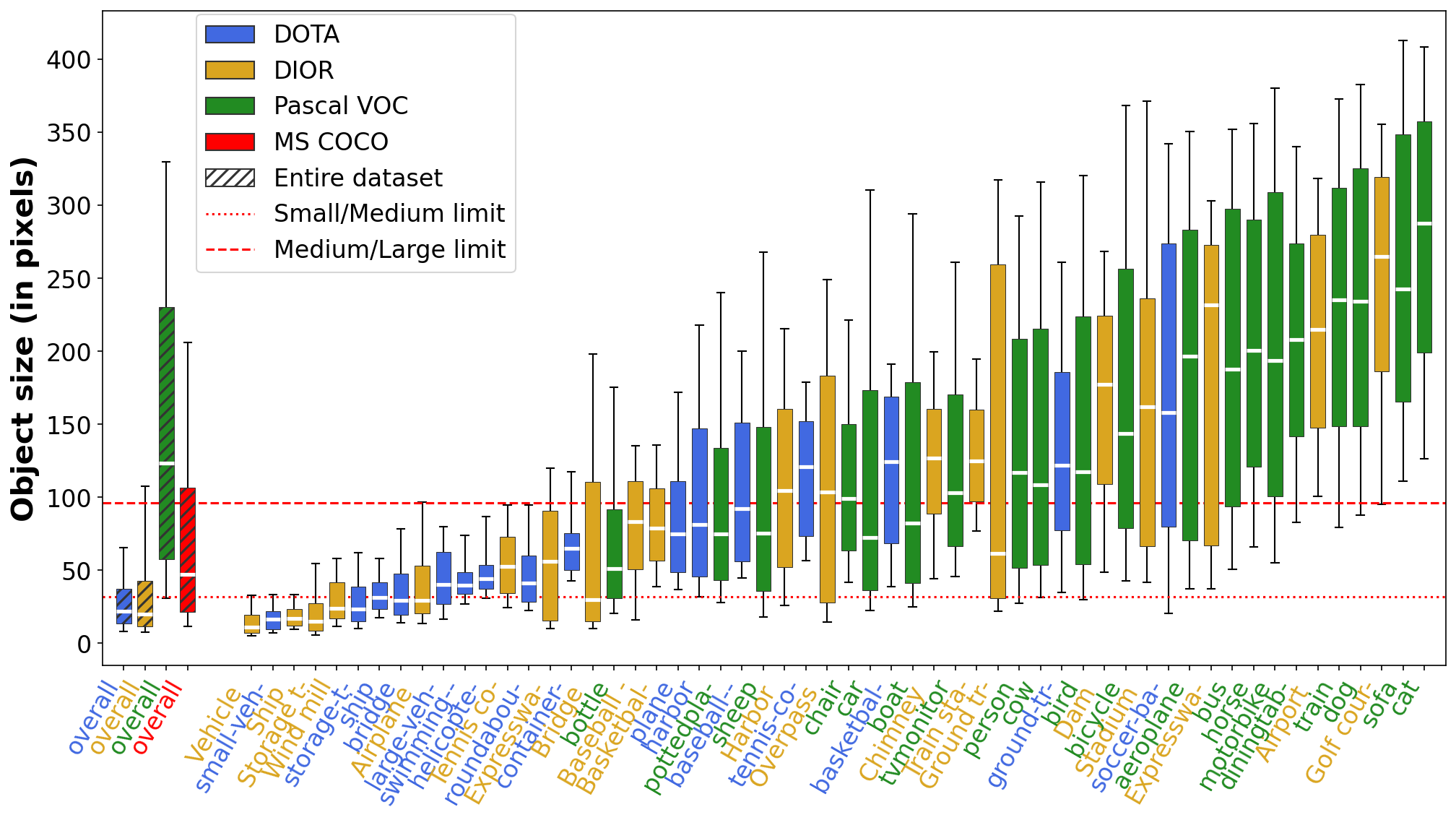}
    \vspace{-5mm}
    \caption{Box plot of objects size in DOTA, DIOR and Pascal VOC and MS COCO.
    On the left side, boxes represent the overall size distribution in each
    dataset. On the right side, the distributions are split by class and ordered
    by average size. As MS COCO contains 80 classes, we chose not to include the
    per class box plots for it in this plot.}
    \label{fig:obj_sizes}
    \vspace{-3mm}
\end{figure}

We hypothesize that this performance gap is mainly due to differences in the
objects sizes within the datasets. In aerial images, objects are much smaller in
average. While this is already an issue for object detection, the problem is
amplified for FSOD as it is difficult to extract semantic information from small
objects in support examples. To support our hypothesis, we first conduct a brief
size analysis of the three datasets DOTA, DIOR, Pascal VOC and MS COCO (see
\cref{fig:obj_sizes}). Aerial datasets contain far smaller objects than natural
ones. Plus, in aerial datasets the size of objects in different classes differs
a lot. Some classes contain only small objects, while others only large objects.
In Pascal VOC, this class' size variance is limited. We argue that it is more
difficult for the model to extract relevant information from small support
examples but also to learn more diverse features to deal with a greater objects'
size variance. This partly explains the greater difficulty of MS COCO.

\begin{figure*}[]
    \centering
    \includegraphics[width=0.97\textwidth]{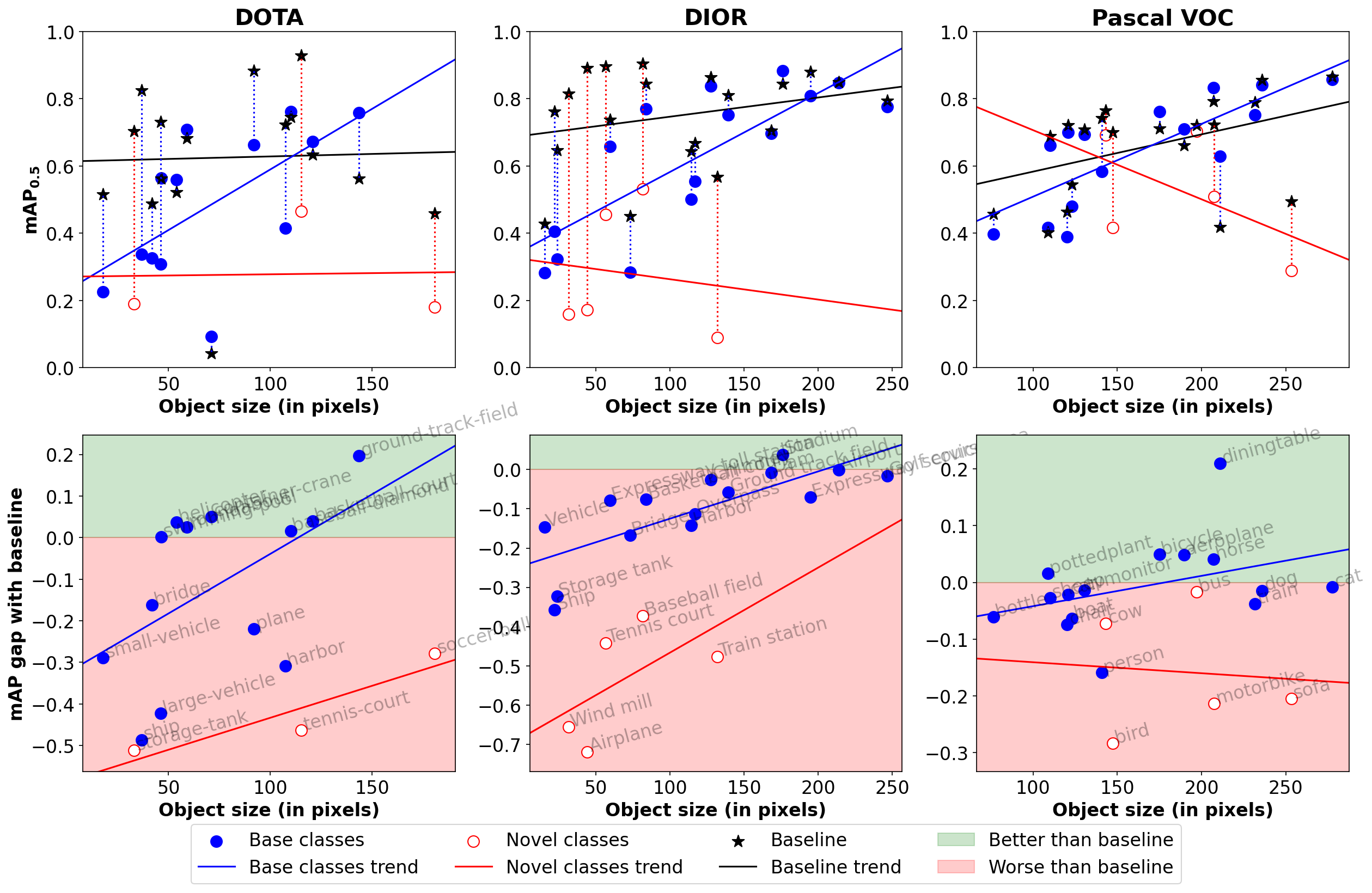}
    \caption{Performance comparison between FRW baseline -- with 10 shots -- (blue and red dots) and
     regular baseline (black stars) on three different datasets: DOTA, DIOR and
     Pascal VOC. \textbf{(top)} Mean average performance of the two methods
     plotted per class against average object size. \textbf{(bottom)} gap
     between FRW baseline and regular baseline, per class. Positive values
     indicate better performance than regular baseline.}
    \label{fig:perf_by_class}
\end{figure*}

To support this claim, we conduct a per-class performance analysis on DOTA, DIOR
and Pascal VOC. The results of this comparison are available in
\cref{fig:perf_by_class}. In this figure, the performance is reported per class
against the average size of the class. The first row reports absolute mAP values
both for FRW and FCOS (baseline). In the second, the mAP gap between the FRW and
the baseline is plotted against the objects' size. We did not report RmAP values
for the sake of visualization. RmAP can take large value (e.g. when baseline
mAP is low) and this squeezes the interesting part of the plot in a
narrow band around 0. The same plot with RmAP is available in Appendix
\ref{app:rmap}. Larger objects are easier to detect, even without
few-shot. This trend is accentuated in the few-shot regime (in the first row,
the blue trend lines are steeper than the blacks). This is observed for base classes
but not always for novel classes, probably because the trends on novel
classes are not reliable due to the limited number of points.
\cref{fig:rmap_all} in Appendix \ref{app:rmap} shows a more reliable trend
for novel classes when the results from the three datasets are aggregated.
Finally, the few-shot inference, which leverage support information to condition
the detection can surpass the baseline in some cases. For base classes, the
model benefits from having examples available at test time only when the objects
are large. On the contrary, when the objects are small, this inference mode is
detrimental. For novel classes, however, the performance is always below the
baseline, even if the gap shrinks with larger objects. This is
expected as the network only received a weak supervision for these classes. 

This comparative analysis confirms that detecting small objects is a very
difficult task in the few-shot regime. It is hard to extract useful information
from small support objects. Even worse, this information can be detrimental for
the detection. Existing FSOD methods are not designed to deal with small
objects, hence the application of these methods on aerial images does not yield
satisfactory results. It is therefore crucial to develop FSOD techniques that
target specifically small objects. We propose a first attempt in this direction
in \cref{sec:xscale}.

\section{Cross-Scales Query-Support Alignment for Small FSOD}
\label{sec:xscale}
From the analysis in \cref{sec:res_aerial}, it is clear that a new attention
mechanism specifically designed for small objects is required to get reasonable
performance on aerial images. To this end, we propose a novel alignment method
that combines information from multiple scales: Cross-Scales Query-Support
Alignment (XQSA). This differs from existing methods which often work
independently at different scales. First, all query features (i.e. from
different levels) are flattened and concatenated. Then, they are linearly
projected into the queries, keys and values matrices $Q$, $K$ and $V$:  



\vspace{6mm}
\begin{align}
    Q = \tikzmarknode{g}{\highlight{Brown}{$F_q$}} W_Q &= [\tikzmarknode{s}{\highlight{Violet}{$F_{q,0}, F_{q,1}, F_{q,2}$}}] \tikzmarknode{n}{\highlight{Green}{$W_Q$}}, \\
    K^c = \tikzmarknode{r}{\highlight{Brown}{$F_s^c$}} W_K &= [\tikzmarknode{l}{\highlight{Violet}{$F_{s,0}^{c}, F_{s,1}^{c}, F_{s,2}^{c}$}}] \tikzmarknode{n}{\highlight{Green}{$W_K$}}, \\
    V^c = \tikzmarknode{r}{\highlight{Brown}{$F_s^c$}} W_V &= [\tikzmarknode{l}{\highlight{Violet}{$F_{s,0}^{c}, F_{s,1}^{c}, F_{s,2}^{c}$}}] \tikzmarknode{t}{\highlight{Green}{$W_V$}}.
\end{align}
\begin{tikzpicture}[overlay,remember picture,>=stealth,nodes={align=left,inner ysep=1pt},<-]
    \path (s.north) ++ (0,1em) node[anchor=south west,color=Violet!80] (lambda){\footnotesize Per level features};
    \draw [color=Violet!80](s.north) |- ([xshift=-0.3ex,color=Violet]lambda.north east);
    \path (g.north) ++ (0,1em) node[anchor=south east,color=Brown!80] (gfeat){\footnotesize Concatenated \\\footnotesize multiscale features};
    \draw [color=Brown!80](g.north) |- ([xshift=-0.3ex,color=Brown]gfeat.north west);
    \path (t.south) ++ (0,-1em) node[anchor=north east,color=Green!80] (mat){\footnotesize Learned projection matrices};
    \draw [color=Green!80](t.south) |- ([xshift=-0.3ex,color=Green]mat.south west);
\end{tikzpicture}
\vspace{3mm}

From this, an affinity matrix is computed between the queries and the keys, and then
the aligned support features are computed as: 

\begin{align}
    \lambda^c &= \text{Softmax}(\frac{Q{K^c}^T}{\sqrt{d}}), \\
    A_q^c &= \lambda^c V^c.
\end{align}

\begin{figure*}
    \centering
    \includegraphics[width=\textwidth]{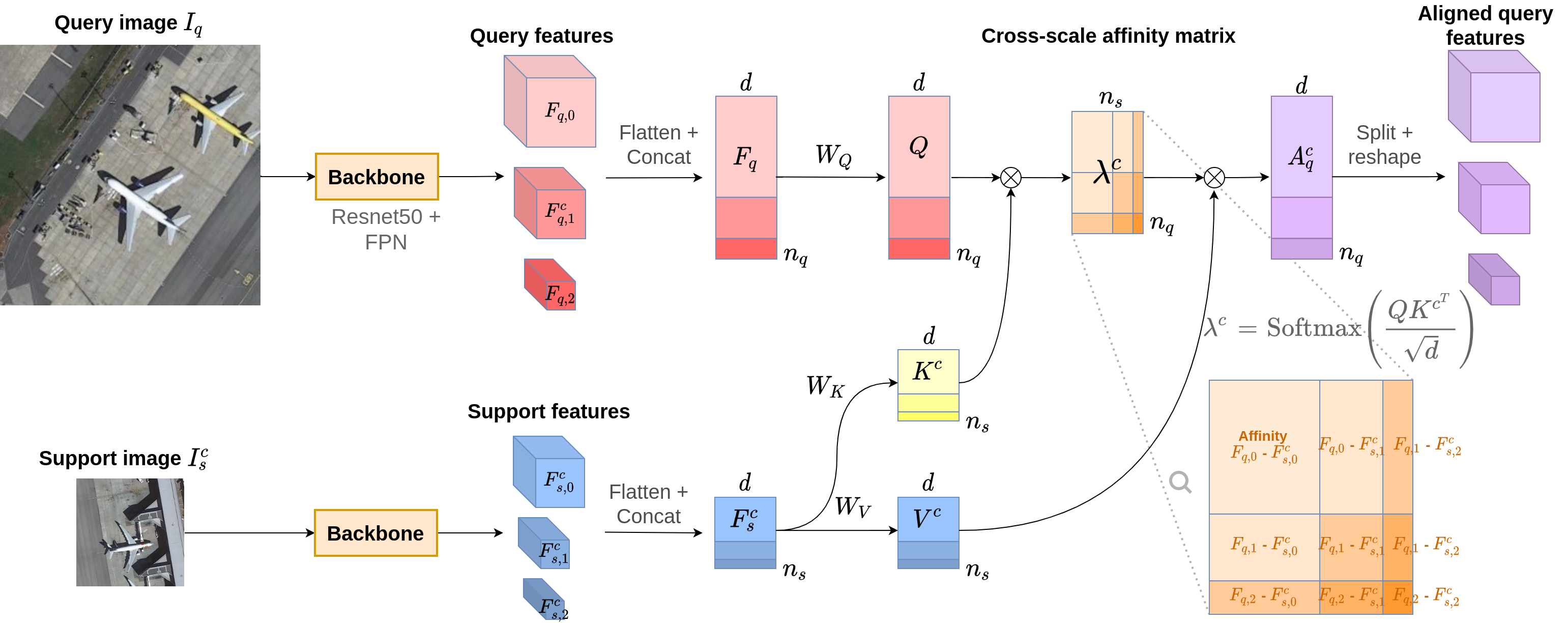}
    \caption{Diagram illustrating the cross-scales query-support alignment
    method. Features are extracted from the query and support images at multiple
    scales and combined to form an affinity matrix. For each query feature
    position, the affinity is computed with any position in the support
    features. This allows object matching across different feature levels.}
    \label{fig:xscale}
\end{figure*}

The aligned features are finally processed by a two-layers MLP with skip
connections. LayerNorm \cite{ba2016layer} is applied before alignment and the
MLP. It can be seen as a learnable fusion operation in the AAF framework,
similar to what was proposed in \cite{li2020one, han2021meta,wu2021universal}
(see \cref{tab:comparison}). This resembles the ViT attention, but with a major
difference, it combines features from different images and different levels (see
\cref{fig:xscale}). This allows to better match objects of different sizes in
the support and the query images.

Small objects have a limited footprint in feature maps which make them hard to
detect but also hard to match with support examples. XQSA's multiscale
alignment enhances the chances of matching as each query feature is compared with
support features at all scales. 
Finally, in order to fairly compare with DANA, we leverage their Background
Attenuation block (BGA) on the support features before alignment. They conduct a
thorough ablation study which shows the positive impact of BGA on the few-shot
performance of their method. We also carry out an ablation study about our
cross-scales method (see Appendix \ref{app:ablation}) and find that BGA also
improves performance in this case. XQSA is implemented inside the AAF framework,
split into three modules: alignment, attention and fusion, following the
description from \cref{sec:framework}. 

To assess the capabilities of the proposed method, we compare it with the best
methods from \cref{sec:res_aerial}: FRW and DANA on DOTA, DIOR and Pascal VOC
(see Appendix \ref{app:res_coco} for results with $\text{mAP}_{0.5:0.95}$ and on
MS COCO). The results of these experiments are available in
\cref{tab:xscale_res}. The mAP values are reported separately for small
($\sqrt{wh} < 32$), medium ($32 \leq \sqrt{wh} < 96$) and large ($\sqrt{wh} \geq
96$) objects. Hence, the methods can be compared specifically on small objects.
XQSA performs consistently better on small and medium novel objects, compared
with FRW and DANA. This is mitigated on base classes, but it is not surprising,
the large number of available examples is enough to learn a robust query-support
matching even for small objects. In some cases, it is worse than the other
methods for base classes. While allowing matching across different scales, XQSA
also brings more mismatches. For base classes, with strong supervision, the
benefits of the multiscale matching may not compensate the drawbacks associated.
On large objects, the performance is a bit lower with XQSA. Probably for the
same reasons as for base classes, large objects are already well handled by
existing FSOD techniques and in this case the multiscale matching is not worth
it. Looking at the performance on all objects, the proposed alignment technique
improves significantly the detection quality for aerial images. Using XQSA in
the AAF framework increased novel class mAP by 0.05 on DOTA and 0.04 on DIOR. 
As it works better on small objects but worse on large objects, it is less
appropriate for natural images. As a consequence, it shows lower improvements
for Pascal VOC and MS COCO. 

Overall, the proposed method largely improves on existing works for aerial
images. On DIOR, this corresponds to a 0.1 mAP increase compared to previous
state-of-the-art \cite{deng2020few}. However, this is not sufficient to fill the
performance gap with natural images as presented in Appendix \ref{app:rmap}.
While XQSA improves on other methods on aerial images, it is still far behind
the performance of the baseline without few-shot. XQSA is better for small
objects but at the cost of lower performance on large objects and base classes.
Progresses are still required to get more versatile FSOD solutions able to
handle small, medium, and large objects.

\begin{table*}[]
    \centering
    \caption{Comparison of the performance between the proposed XQSA alignment
    method and the two best reimplemented methods on DOTA and DIOR, FRW and
    DANA. $\text{mAP}_{0.5}$ values are reported separately for base and novel
    classes on DOTA, DIOR and Pascal VOC. Results with $\text{mAP}_{0.5}$ and on
    MS COCO are also available in Appendix \ref{app:res_coco}. As XQSA is
    specifically designed for small objects, mAP values are reported for All,
    Small ($\sqrt{wh} < 32$), Medium ($32 \leq \sqrt{wh} < 96$) and Large
    ($\sqrt{wh} \geq 96$) objects. All experiments were conducted with $K=10$
    shots.}
    \label{tab:xscale_res}
    \resizebox{\textwidth}{!}{%
    \begin{tabular}{@{}lcrrrlrrrrccrrrlrrrrccrrrlrrrr@{}}
        \toprule[1pt]
        \multicolumn{1}{c}{}          & \multicolumn{9}{c}{\textbf{DOTA}}                                                                                                                                                                                                                                             & \textbf{}            & \multicolumn{9}{c}{\textbf{DIOR}}                                                                                                                                                                                                                                             & \textbf{}            & \multicolumn{9}{c}{\textbf{Pascal VOC}}                                                                                                                                                                                                                                       \\ \cmidrule(lr){2-10} \cmidrule(lr){12-20} \cmidrule(l){22-30} 
        \multicolumn{1}{c}{\textbf{}} & \multicolumn{4}{c}{\textbf{Base}}                                                                                                    &  & \multicolumn{4}{c}{\textbf{Novel}}                                                                                                  & \textbf{}            & \multicolumn{4}{c}{\textbf{Base}}                                                                                                    &  & \multicolumn{4}{c}{\textbf{Novel}}                                                                                                  & \textbf{}            & \multicolumn{4}{c}{\textbf{Base}}                                                                                                    &  & \multicolumn{4}{c}{\textbf{Novel}}                                                                                                  \\ \midrule
        \multicolumn{1}{c}{\textbf{}} & \textbf{All}                      & \multicolumn{1}{c}{\textbf{S}} & \multicolumn{1}{c}{\textbf{M}} & \multicolumn{1}{c}{\textbf{L}} &  & \multicolumn{1}{c}{\textbf{All}} & \multicolumn{1}{c}{\textbf{S}} & \multicolumn{1}{c}{\textbf{M}} & \multicolumn{1}{c}{\textbf{L}} & \textbf{}            & \textbf{All}                      & \multicolumn{1}{c}{\textbf{S}} & \multicolumn{1}{c}{\textbf{M}} & \multicolumn{1}{c}{\textbf{L}} &  & \multicolumn{1}{c}{\textbf{All}} & \multicolumn{1}{c}{\textbf{S}} & \multicolumn{1}{c}{\textbf{M}} & \multicolumn{1}{c}{\textbf{L}} & \textbf{}            & \textbf{All}                      & \multicolumn{1}{c}{\textbf{S}} & \multicolumn{1}{c}{\textbf{M}} & \multicolumn{1}{c}{\textbf{L}} &  & \multicolumn{1}{c}{\textbf{All}} & \multicolumn{1}{c}{\textbf{S}} & \multicolumn{1}{c}{\textbf{M}} & \multicolumn{1}{c}{\textbf{L}} \\
        \textbf{FRW}                  & \multicolumn{1}{r}{0.49}          & 0.25                           & 0.59                           & 0.63                           &  & 0.37                             & 0.14                           & 0.34                           & 0.59                           & \multicolumn{1}{l}{} & \multicolumn{1}{r}{0.62}          & 0.08                           & \textbf{0.49}                  & 0.81                           &  & 0.36                             & 0.02                           & 0.34                           & 0.59                           & \multicolumn{1}{l}{} & \multicolumn{1}{r}{0.63}          & 0.16                           & 0.48                           & \textbf{0.82}                  &  & 0.49                             & 0.16                           & 0.27                           & \textbf{0.68}                  \\
        \textbf{DANA}                 & \multicolumn{1}{r}{\textbf{0.54}} & \textbf{0.37}                  & \textbf{0.62}                  & \textbf{0.70}                  &  & 0.36                             & 0.14                           & 0.40                           & \textbf{0.65}                  & \multicolumn{1}{l}{} & \multicolumn{1}{r}{\textbf{0.63}} & \textbf{0.11}                  & \textbf{0.49}                  & \textbf{0.83}                  &  & 0.38                             & 0.03                           & 0.35                           & \textbf{0.61}                  & \multicolumn{1}{l}{} & \multicolumn{1}{r}{\textbf{0.65}} & \textbf{0.18}                  & \textbf{0.51}                  & 0.80                           &  & 0.52                             & 0.10                           & 0.25                           & 0.67                           \\
        \textbf{XQSA}              & \multicolumn{1}{r}{0.51}          & 0.26                           & 0.59                           & 0.64                           &  & \textbf{0.41}                    & \textbf{0.18}                  & \textbf{0.45}                  & 0.54                           & \multicolumn{1}{r}{} & \multicolumn{1}{r}{0.60}          & \textbf{0.11}                  & 0.46                           & 0.82                           &  & \textbf{0.42}                    & \textbf{0.04}                  & \textbf{0.41}                  & 0.58                           & \multicolumn{1}{r}{} & \multicolumn{1}{r}{0.62}          & 0.16                           & 0.49                           & 0.76                           &  & \textbf{0.54}                    & \textbf{0.19}                  & \textbf{0.35}                  & 0.66                           \\ \bottomrule[1pt]
        \end{tabular}%
        }
\end{table*}

\section{Conclusion and Future Work}
\label{sec:conclusion}
In a nutshell, our review of attention-based FSOD emphasizes the variety of
methods in the literature and the lack of application on aerial images.
Comparison between FSOD methods is difficult as many details change from a
method to another. This makes it impractical to find the most suitable method
for a given problem or dataset. Our proposed framework solves this issue by
providing a fixed but modular environment to benchmark attention-based FSOD
methods. Using this framework, we conduct several comparisons on various
datasets. We also provide a novel metric, Relative mAP (RmAP), specifically
designed to assess how good a FSOD method is compared with its non few-shot
counterpart. This metric allows comparison between different datasets, which
highlights different behaviors for FSOD methods on natural and
aerial images. Our experiments provide evidences toward the nature of these
discrepancies. Small objects are notoriously difficult to detect, but in the
few-shot regime they are even more challenging as the query-support matching
is harder for them. To close the performance gap between small, medium and large
objects, we propose a novel multiscale alignment method named XQSA. It is designed to
enlarge the matching possibilities across different feature levels. This improves
significantly the performance on small and medium objects, yielding large
improvements over the state-of-the-art on DOTA and DIOR datasets. However, this
comes at the expense of lower performance on large objects and base classes.
Future works should thus focus on developing more versatile FSOD techniques
able to deal both with small and large objects. This work shed some light on the
challenging small few-shot object detection problem and provide tools for deeper
analysis and the development of more universal attention based FSOD methods.


\ifCLASSOPTIONcompsoc
  \section*{Acknowledgments}
\else
  \section*{Acknowledgment}
\fi
The authors would like to thank COSE company for their close collaboration and the
funding of this project.

\ifCLASSOPTIONcaptionsoff
  \newpage
\fi

\bibliographystyle{IEEEtran}
\bibliography{IEEEabrv,bibliography}

\vspace{-10mm}

%
\begin{IEEEbiography}[{\includegraphics[width=1in,height=1.25in,clip,keepaspectratio]{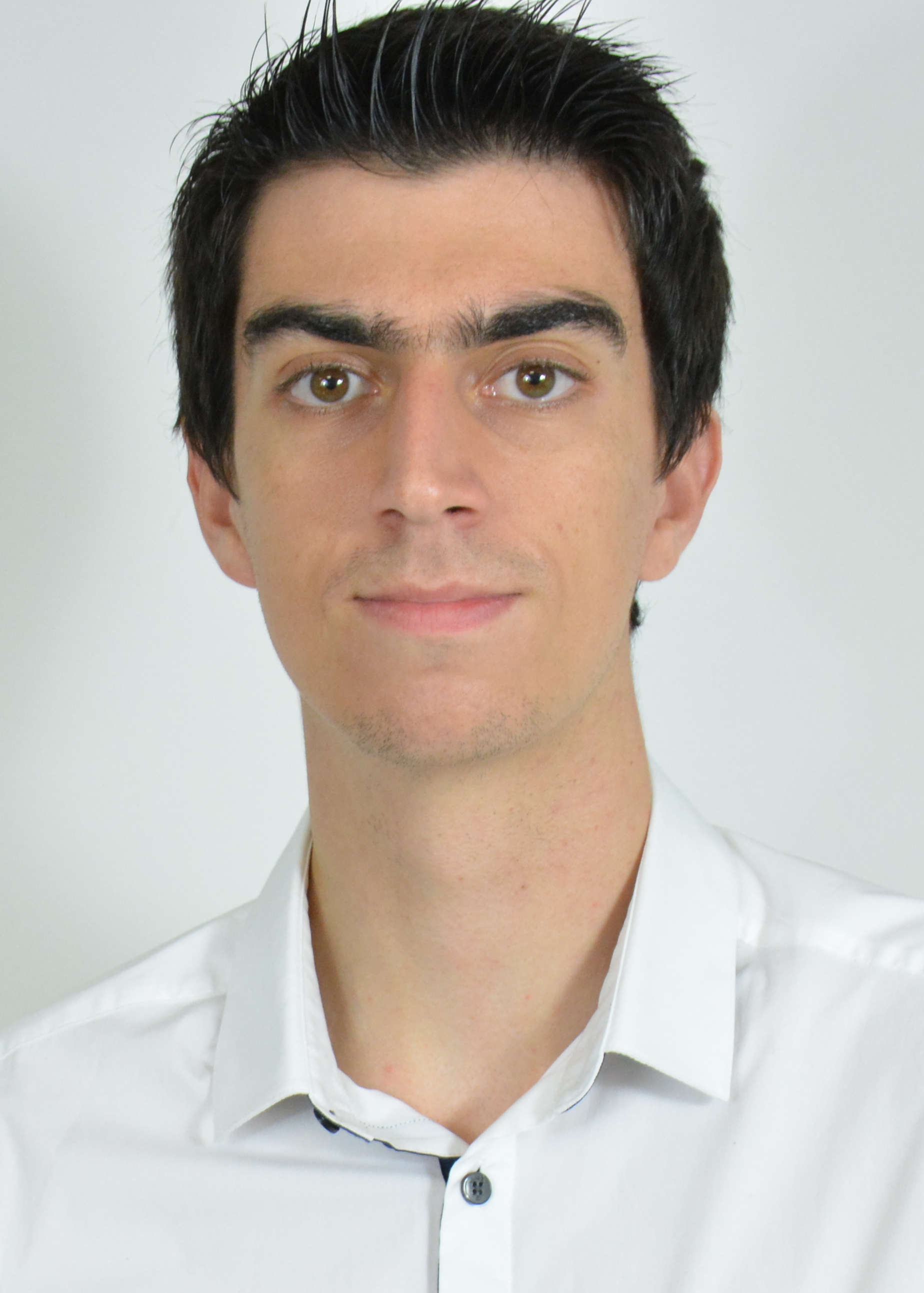}}]{Pierre LE JEUNE}
  is a PhD student at L2TI laboratory, University Sorbonne Paris Nord while
  working at COSE company. He received the M.Sc. degree in Mathematical Modelling
  and Computation from Danish Technical University and the M.Sc. in Engineering
  from Centrale Nantes. His current research interests include Few-Shot Learning,
  Computer Vision and Deep Learning.
\end{IEEEbiography}
\vspace{-10mm}
\begin{IEEEbiography}[{\includegraphics[width=1in,height=1.25in,clip,keepaspectratio]{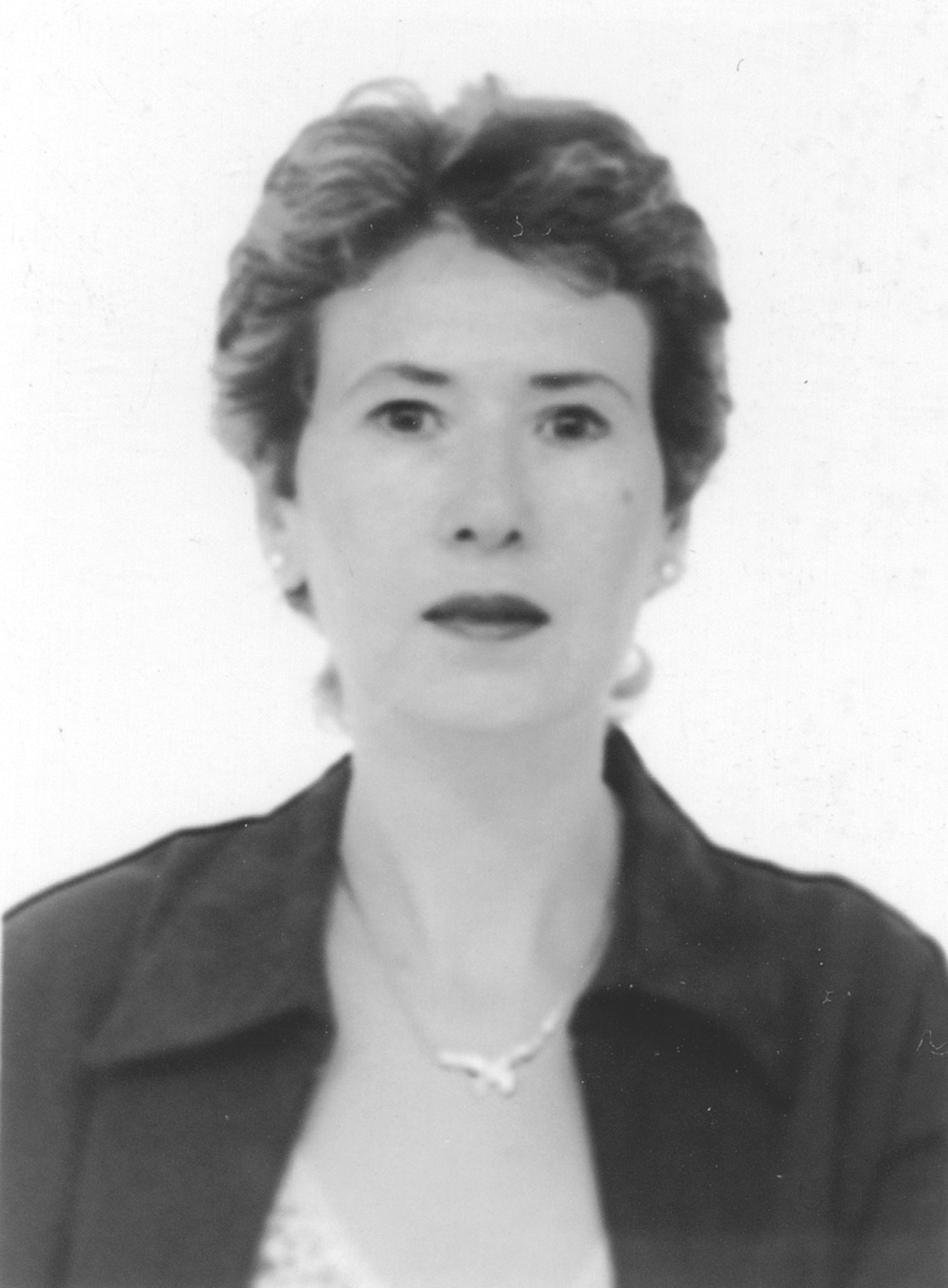}}]{Anissa MOKRAOUI}
  received the state engineering degree in electrical
  engineering from national school of telecommunications in 1989 from Algeria,
  the M.S degree in information technology in 1990 and the Ph.D. degree in 1994
  both from University Paris 11, Orsay France. From 92 to 94, she worked at the
  National Institute of Telecommunications (INT, at Evry France) where her
  research activities were on digital signal processing, fast filtering
  algorithms and implementation problems on DSP. In 1997, she was appointed as
  assistant professor and in December 2011 as associate professor at Galil\'e
  Institute of University Paris 13, France. Since 2013, she is full professor at
  Galil\'ee Institute of Universit\'e  Sorbonne Paris Nord (USPN), France. Since
  2016, she is the director of the L2TI laboratory of USPN. Her current research
  interests include source coding (image, video, multi-view, stereoscopic); joint
  source-channel-protocol decoding, robust mobile transmission, MIMO-OFDM
  channel estimation (massive), computer vision, few-shot object detection. She
  is co-author of more than 150 contributions to journals and conference
  proceedings. She served on program committees for conferences. She acts as a
  reviewer for many IEEE and EURASIP conferences and journals.
\end{IEEEbiography}

\clearpage
\appendices

\section{Novel classes splits}
\label{app:class_split}

The base/novel classes split is crucial to compare the performance reported for
FSOD methods. This appendix reports the splits used in our experiments. The
human-readable labels are not included in the following table to keep the
size constrained, but they can be found on datasets' project pages.   

\begin{table}[h]
    \centering
    \caption{Base / Novel class splits for the different datasets used in this work.}
    \label{tab:class_split}
    \resizebox{\columnwidth}{!}{%
    \begin{tabular}{@{}lll@{}}
    \toprule
                        & \textbf{Novel classes}                              & \textbf{Base classes}                                                                                                                                                              \\ \midrule 
    \makecell[b]{\textbf{Pascal VOC}} & 3,6,10,14,18                                        & 1,2,4,5,7,8,9,11,12,13,15,16,17,19,20                                                                                                                                              \\ \addlinespace[0.5em]
    \makecell[t]{\\\\\textbf{MS COCO}   } & \makecell[lt]{1,2,3,4,5, \\ 6,7,9,15,16,\\17,18,19,20,40,\\57,58,59,61,63} & \makecell[lt]{8,10,11,12,13,14,21,22,23,24,\\25,26,27,28,29,30,31,32,33,34,\\35,36,37,38,39,41,42,43,44,45,\\46,47,48,49,50,51,52,53,54,55,\\56,60,62,64,65,66,67,68,69,70,\\71,72,73,74,75,76,77,78,79,80} \\ \addlinespace[0.5em]
    \makecell[b]{\textbf{DOTA}      } & 3,5,15                                            & 1,2,4,6,7,8,9,10,11,12,13,14,16                                                                                                                                                    \\ \addlinespace[0.5em]
    \makecell[b]{\textbf{DIOR}      } & 1,3,17,18,20                                        & 2,4,5,6,7,8,9,10,11,12,13,14,15,16,19                                                                                                                                              \\ \bottomrule
    \end{tabular}%
    }
    \end{table}

\section{Object-level augmentations}
\label{app:augmentations}

To improve the performance of the methods implemented in the AAF framework and
be competitive with existing works, we propose an augmentation pipeline specifically
designed for detection. Some regular augmentation techniques cannot be directly
applied for object detection as it can completely mask objects from the image.
This deteriorates the training as the model will not be able to detect hidden
objects, but it will be penalized anyway. 

\begin{figure}[h]
    \centering
    \includegraphics[width=\columnwidth]{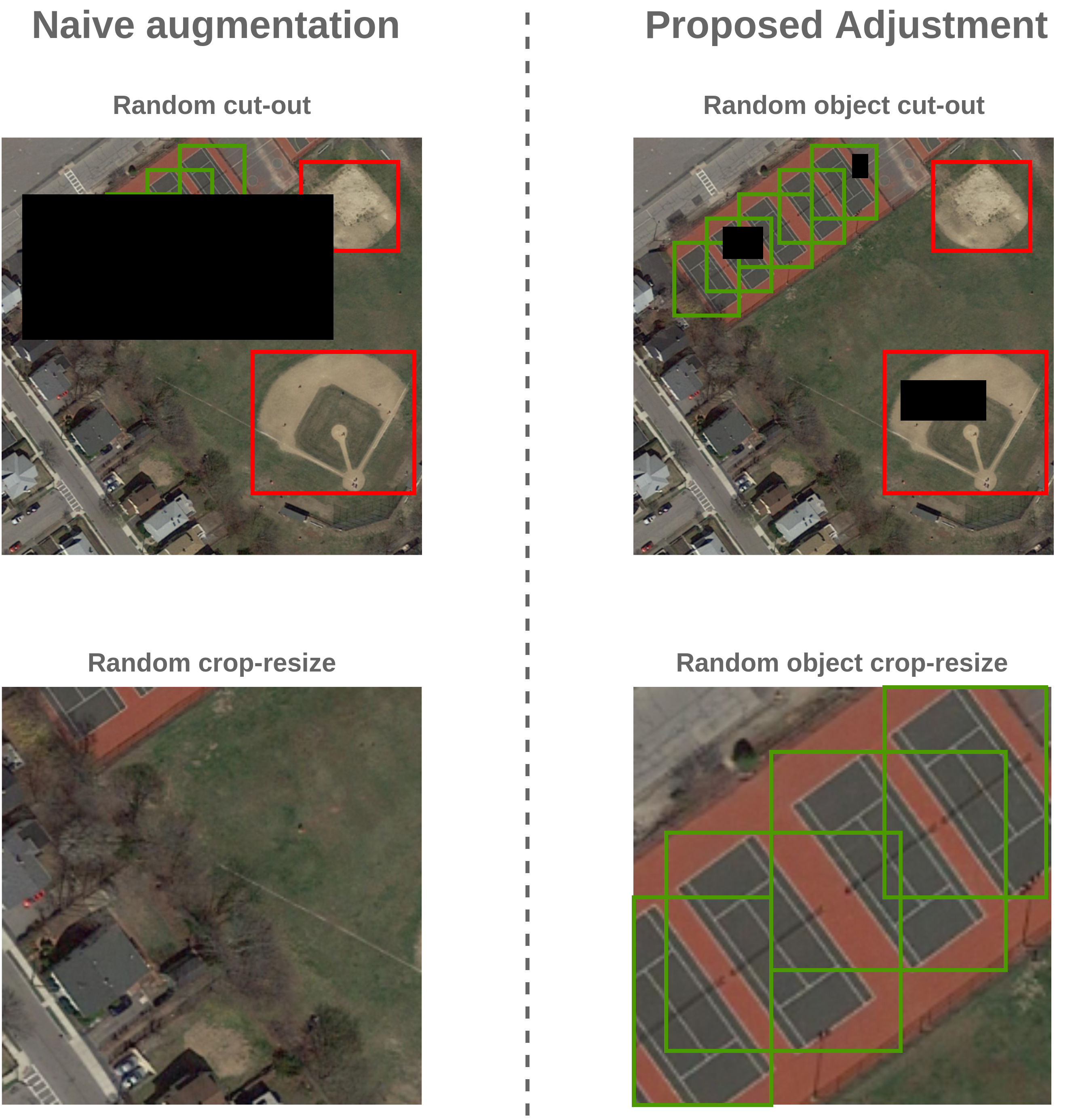}
    \caption{Difference between naive augmentation techniques (left) and our
    adaptation to object detection (right). The proposed transformations are
    applied at the object-level to preserve objects integrity.}
    \label{fig:augmentations}
\end{figure}

First, we apply random horizontal and vertical flips (only for aerial images)
and color jitter. As it does not remove entire objects, these can be applied
directly on the images. Unlike some other classical techniques such as random
crop-resize and random cut-out cannot. Therefore, we developed
object-preserving random crop-resize and cut-out. The main idea is
to apply these transformations at the object level and not at the image level.
This ensures that objects of interest are still visible in the transformed
image. For crop-resize, a non-empty subset of the objects in the image is
randomly sampled. An overall bounding box is computed around all these objects
and the cropped area is randomly drawn between this box and the image borders.
Hence, it guarantees the presence of at least one object inside the cropped
image. For cut-out, the principle is similar, instead of cutting out a random
part of the image, the cut is applied at the object level so that it does not
hide out entire objects. \cref{fig:augmentations} compares the two
proposed augmentations with their naive implementations. 

We performed a cumulative study to assess the benefits of each component of the
augmentation pipeline. This is summarized in
\cref{tab:augmentation-performance}. It shows that the augmentation is
beneficial for the performance on novel classes but detrimental for base
classes. It is surprising that performance drops on base classes with
augmentation. 
More specifically, it seems that image flips are responsible for the performance
loss on base classes (see first and second columns in
\cref{tab:augmentation-performance}). Base classes performance drops when adding
flips but remains mostly constant when adding other types of augmentations. One
crucial difference between flip and other augmentations is that we chose to
apply flips also on support examples. This choice was made to increase the
diversity of the support set during fine-tuning. For novel classes, only a few
images are available as support during fine-tuning, and we want to avoid
overfitting these examples. Although other types of augmentation could have been
employed for this as well, we wanted to prevent disrupting too much the
information in the support. This choice may be the cause of the performance drop
on base classes. To verify this hypothesis, we conduct a few more experiments
disabling the flip in the support set. With the \textit{default} cropping
strategy, the experiments confirm the hypothesis: no performance drop is
observed when supports are not flipped. However, as this certainly interacts
with the cropping strategy, we also tried with \textit{same-size} as it is the
chosen strategy for most experiments in this paper (see Appendix
\ref{app:cropping}). Surprisingly, it does not produce similar results, and in
this case, flips in supports are actually beneficial for base classes
performance. This suggests a complex interaction between augmentation on the
support set and the cropping strategy. The choice made in this paper may not be
optimal in this regard, and a deeper analysis of this interaction should be
conducted as future work. Finally, the base class performance loss is
compensated by clear improvements on novel classes. As this is the main goal of
FSOD, we chose to adopt the original augmentation pipeline, i.e. with flips in
support, for our experiments.  

\begin{table}[h]
    \centering
    \caption{Cumulative study of the proposed augmentation techniques on DOTA
    with FRW method. $\text{mAP}_{0.5}$ is reported for different
    number of shots.}
    \label{tab:augmentation-performance}
    \resizebox{0.45\textwidth}{!}{%
    \begin{tabular}{@{}lllllll@{}}
        \toprule
        \textbf{\# Shots}   & \textbf{}      & \textbf{Baseline} & \textbf{+ Flip} & \textbf{+ Color} & \textbf{+ Cutout} & \textbf{+ Crop} \\ \midrule
        \multirow{2}{*}{1}  & \textbf{Base}  & \textbf{0.488}    & 0.458            & 0.460            & 0.472             & 0.457           \\
                            & \textbf{Novel} & 0.062             & 0.052            & 0.069            & 0.064             & \textbf{0.100}  \\ \addlinespace[1mm]
        \multirow{2}{*}{3}  & \textbf{Base}  & \textbf{0.511}    & 0.475            & 0.470            & 0.461             & 0.452           \\
                            & \textbf{Novel} & 0.144             & 0.186            & 0.186            & 0.197             & \textbf{0.220}  \\ \addlinespace[1mm]
        \multirow{2}{*}{5}  & \textbf{Base}  & \textbf{0.527}    & 0.494            & 0.501            & 0.503             & 0.487           \\
                            & \textbf{Novel} & 0.193             & 0.237            & 0.251            & 0.250             & \textbf{0.259}  \\ \addlinespace[1mm]
        \multirow{2}{*}{10} & \textbf{Base}  & \textbf{0.538}    & 0.508            & 0.508            & 0.504             & 0.503           \\
                            & \textbf{Novel} & 0.286             & 0.312            & 0.281            & 0.341             & \textbf{0.359}  \\ \bottomrule
        \end{tabular}
    }
    \vspace{-4mm}
    \end{table}

\section{Support extraction strategies}
\label{app:cropping}
The support information is located only inside a delimited area of the support
image, its corresponding bounding box. The remaining part of the image mostly
contains irrelevant information concerning the object class. Therefore,
extracting features from the whole support images is not necessary. But features
contained only inside the object's bounding box might not be sufficient as well.
Context can be extremely valuable in certain cases, especially for small
objects. For instance, a car and a small boat could easily be mistaken without
context. Close surroundings of the objects can help for recognition. 

The most common strategy for support information extraction is proposed by
\cite{kang2019few}. They concatenate the whole support image with the support
object's binary mask (rectangular, computed from the bounding box) and pass this
to an extractor network. This has two main drawbacks. First, it is necessary to
compute feature from the entire support image, which is a loss of resources.
Second, the same network cannot be used for extracting features in query and
support images as it needs an additional input channel to process the mask.
Hence, the network cannot be pretrained beforehand. This design choice is rarely
discussed, if ever mentioned, in the literature.

\begin{figure}[!ht]
    \centering
    \includegraphics[width=0.95\columnwidth]{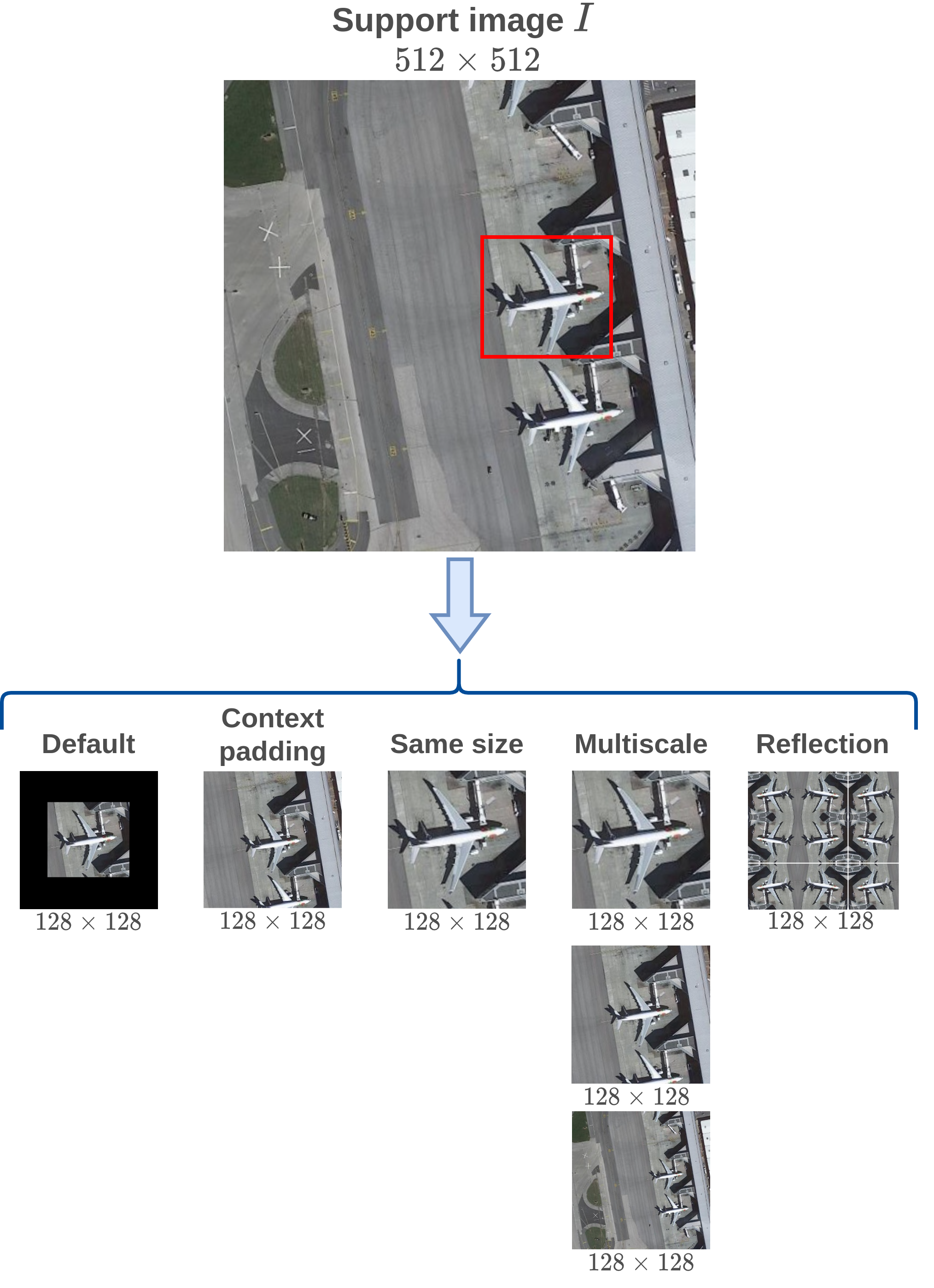}
    \caption{Illustration of the different cropping strategies tested. The
    mixed strategy is not illustrated as it is a combination of \textit{default}
    and \textit{same-size}.}
    \label{fig:cropping}
\end{figure}

In this appendix, we explore this design choice by implementing several
extraction strategies. We did not reimplement the technique from
\cite{kang2019few} as it requires to have two different networks for support and
query feature extraction. However, some of our techniques are quite close from
what they proposed. These techniques are described bellow and
\cref{fig:cropping} illustrates most of them.

\begin{itemize}
    \item[-]  \textbf{Default}: the most naive extraction technique. It consists in
    cropping the support image around the support object at a fixed size (e.g. $128\times128$).
    Objects larger than this are simply resized to fit in the patch.
    \item[-]  \textbf{Context-padding}: the cropping occurs exactly as with the
    default strategy, but areas around the objects are masked out. This is close
    to what was proposed by \cite{kang2019few}.
    \item[-]  \textbf{Reflection}: context is replaced by reflection of the
    object. In the case of small objects, the support patch can easily be dominated either
    by irrelevant information or by zeros when using the latter two extraction methods.
    \item[-]  \textbf{Same-size}: all objects are resized to fill entirely the
    support patch (preserving the aspect ratio). It does not change the process
    for large objects, but it prevents smaller objects to be dominated by
    irrelevant information.
    \item[-] \textbf{Multi-scale}: the object is resized and cropped at 3
    different scales. Each scale is responsible for matching small, medium and
    large objects in query images.
    \item[-] \textbf{Mixed}: it is a combination of the default strategy and
    \textit{same-size}. Small objects (i.e. $\sqrt{wh} < 32$) are extracted
    using the default strategy. Larger objects ($\sqrt{wh} \geq 32$) are resized
    into a patch of $128 \times 128$ pixels. Hence, small objects are not
    upscaled more than 4 times, as they are using the \textit{resize} strategy.
\end{itemize}

These strategies are compared in Table \ref{tab:cropping_methods}. Even though
\textit{same-size} gets the best overall results on novel classes (regardless of
object sizes), there is no clear conclusion. It is outperformed by both
\textit{reflection} and \textit{mixed} for base classes. No method outperforms
the others on all object sizes, not even the ones designed to be more robust to
size (\textit{multiscale} and \textit{mixed}). The latter two techniques
introduce discrepancies in the features: objects of similar sizes can be
processed differently and give different features. It is probably
easier for the network to learn semantic representations from objects of the
same size. As \textit{same-size} gives the best results on novel classes, we
chose to use this one for all our experiments.

However, in the light of our performance analysis in \cref{sec:res_aerial}, we
can understand some results from \cref{tab:cropping_methods}. The
\textit{multiscale} strategy does not perform very well as it introduces small
objects features which seem detrimental for the good conditioning of the
network. On the contrary, \textit{same-size} only generates large objects as
support which is a better strategy. Finally, \textit{reflection} performs
surprisingly well for small objects while preserving their small size. The
redundancy generated by the reflection of such small objects reinforces the
object's features. 

\begin{figure*}
    \centering
    \includegraphics[width=\textwidth, trim=0 0 0 5,clip]{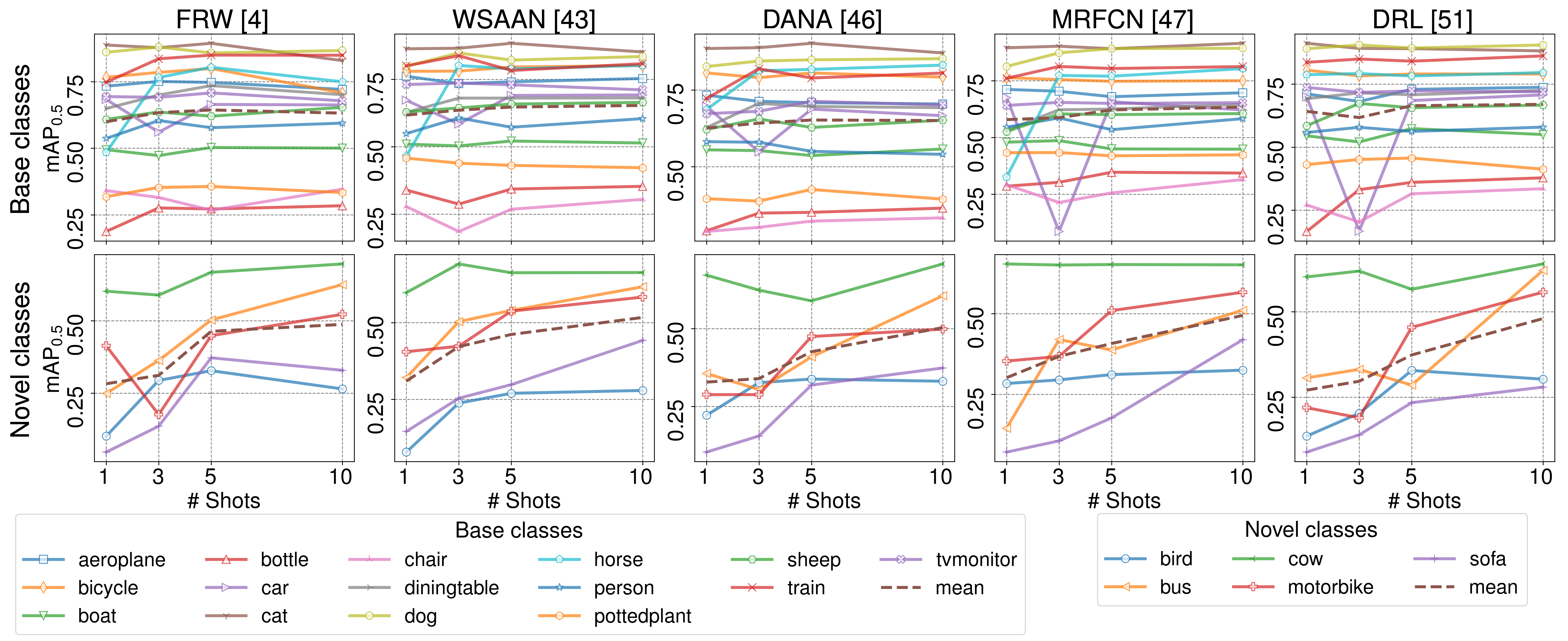}
    \caption{$\text{mAP}_{0.5}$ on Pascal VOC against the number of shots for
            each class and each method. Dashed lines represent average
            performance on all classes, either base classes \textbf{(top)} or novel
            classes \textbf{(bottom)}.}
    \label{fig:map_per_class}
\end{figure*}

\begin{table}[!ht]
    \centering
    \caption{Comparison of support extraction strategies on base and novel
    classes with DOTA dataset and FRW method with 10 shots. $\text{mAP}_{0.5}$
    is reported on all objects and separately on objects of different sizes: small
    (S), medium (M) and large (L).}
    \label{tab:cropping_methods}
    \resizebox{0.5\textwidth}{!}{%
    \begin{tabular}{@{}lcccccccccl@{}}
        \toprule
                                 & \multicolumn{4}{c}{\textbf{Base classes}}                         &  & \multicolumn{4}{c}{\textbf{Novel classes}}                        &  \\ \cmidrule(lr){2-5} \cmidrule(lr){7-10}
                                 & \textbf{All}  & S              & M              & L              &  & \textbf{Mean}  & S              & M              & L              &  \\ \midrule
        \textbf{Default}         & 0.498          & 0.243          & 0.574          & \textbf{0.639} &  & 0.250          & 0.077          & 0.243          & 0.345          &  \\
        \textbf{Context padding} & 0.500          & 0.250          & 0.597          & 0.630          &  & 0.296          & \textbf{0.107} & 0.268          & 0.509          &  \\
        \textbf{Same-size}       & 0.506          & \textbf{0.308} & 0.594          & 0.625          &  & \textbf{0.322} & 0.083          & 0.332          & \textbf{0.565} &  \\
        \textbf{Multiscale}      & \textbf{0.514} & 0.290          & 0.598          & 0.633          &  & 0.270          & 0.084          & \textbf{0.335} & 0.456          &  \\
        \textbf{Reflection}      & 0.503          & 0.261          & 0.594          & 0.625          &  & 0.255          & 0.072          & 0.207          & 0.443          &  \\
        \textbf{Mixed}           & 0.509          & 0.272          & \textbf{0.605} & 0.607          &  & 0.280          & 0.095          & 0.265          & 0.489          &  \\ \bottomrule
        \end{tabular}%
    }
    \vspace{-4mm}
    \end{table}

\section{Results from original papers on Pascal VOC and MS COCO}
\label{app:res_orig}
In this appendix, we report the results on Pascal VOC and MS COCO of the
different methods to be compared with our results achieved in our framework: FRW
\cite{kang2019few}, WSAAN \cite{xiao2020fsod}, DANA \cite{chen2021should}, MFRCN
\cite{han2021meta} and DRL \cite{liu2021dynamic}. These results can be found in
\cref{tab:res_orig_pascal} and \cref{tab:res_orig_coco}. This demonstrates the
lack of comparative study in the literature, most methods are designed and
evaluated with different datasets and metrics. In addition, we indicate in the
table the backbone used by each method to highlights the differences preventing
a good comparison. However, plenty of other details change as well. As an
example, DRL leverages a dynamic relation matrix between support and query
features, solely for the computation of an auxiliary loss function. 

\begin{table}[h]
    \centering
    \caption{$\text{mAP}_{0.5}$ values on Pascal VOC for FRW, WSAAN, DANA, MFRCN
    and DRL as reported in their respective papers. "--" indicates that the
    value is not reported in the paper. }
    \label{tab:res_orig_pascal}
    \resizebox{\columnwidth}{!}{%
    \begin{tabular}{@{}lcccccccccc@{}}
    \toprule[1pt]
    Method & \multicolumn{2}{c}{\textbf{FRW \cite{kang2019few}}}                               & \multicolumn{2}{l}{\textbf{WSAAN \cite{xiao2020fsod}}} & \multicolumn{2}{l}{\textbf{DANA \cite{chen2021should}}}     & \multicolumn{2}{l}{\textbf{MFRCN \cite{han2021meta}}}        & \multicolumn{2}{c}{\textbf{DRL \cite{liu2021dynamic}}}          \\
    Backbone  & \multicolumn{2}{c}{Darknet}                           & \multicolumn{2}{c}{-}     & \multicolumn{2}{c}{ResNet50} & \multicolumn{2}{c}{ResNet101}    & \multicolumn{2}{c}{-}            \\ \midrule
    \# Shots      & \underline{Base}          & \underline{Novel}         & \underline{Base}& \underline{Novel}& \underline{Base}& \underline{Novel}& \underline{Base} & \underline{Novel} & \underline{Base} & \underline{Novel} \\
    1                              & -                         & 0.148                     & -           & -           & -            & -             & -    & 0.430                     & -    & 0.280                     \\
    3                              & -                         & 0.267                     & -           & -           & -            & -             & -    & 0.606                     & -    & 0.494                     \\
    5                              & -                         & 0.339                     & -           & -           & -            & -             & -    & 0.661                     & -    & 0.499                     \\
    10                             & \multicolumn{1}{r}{0.697} & 0.472                     & -           & -           & -            & -             & -    & 0.654                     & -    & 0.594                     \\ \bottomrule[1pt]
    \end{tabular}%
    }
    \end{table}

\begin{table}[h]
    \centering
    \caption{$\text{mAP}_{0.5:0.95}$ values on MS COCO for FRW, WSAAN, DANA, MFRCN
    and DRL as reported in their respective papers. "--" indicates that the
    value is not reported in the paper.}
    \label{tab:res_orig_coco}
    \resizebox{\columnwidth}{!}{%
    \begin{tabular}{@{}lcccccccccc@{}}
    \toprule[1pt]
    Method & \multicolumn{2}{c}{\textbf{FRW \cite{kang2019few}}}                               & \multicolumn{2}{l}{\textbf{WSAAN \cite{xiao2020fsod}}} & \multicolumn{2}{l}{\textbf{DANA \cite{chen2021should}}}     & \multicolumn{2}{l}{\textbf{MFRCN \cite{han2021meta}}}        & \multicolumn{2}{c}{\textbf{DRL \cite{liu2021dynamic}}}          \\
    Backbone  & \multicolumn{2}{c}{Darknet}                           & \multicolumn{2}{c}{-}     & \multicolumn{2}{c}{ResNet50} & \multicolumn{2}{c}{ResNet101}    & \multicolumn{2}{c}{-}            \\ \midrule
    \# Shots      & \underline{Base}          & \underline{Novel}         & \underline{Base}& \underline{Novel}& \underline{Base}& \underline{Novel}& \underline{Base} & \underline{Novel} & \underline{Base} & \underline{Novel} \\
    10              & -            & 0.056          & -           & -           & -    & 0.186                     & -    & 0.127                     & -    & 0.109                     \\
    30              & -            & 0.091          & -           & -           & -    & 0.216                     & -    & 0.166                     & -    & 0.15                      \\ \bottomrule[1pt]
    \end{tabular}%
    }
    \end{table}

\section{Evolution of mAP with the number of shots}
\label{app:map_shot_class}

\begin{figure}[h]
    \centering
    \includegraphics[width=\columnwidth, trim=10 0 0 10,clip]{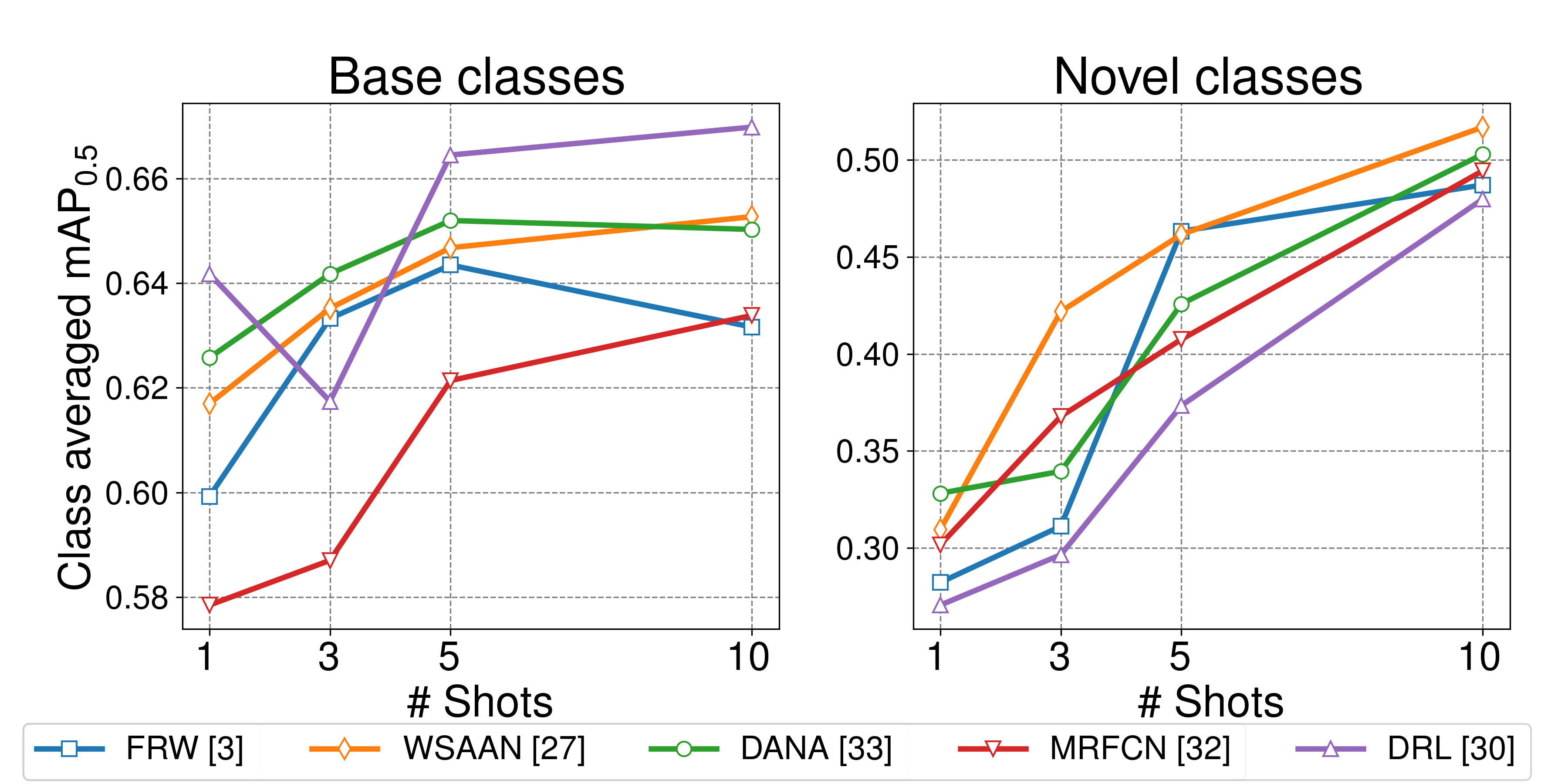}
    \caption{Evolution of $\text{mAP}_{0.5}$ with the number of shots averaged
            on base and novel classes separately. Each line represents one of
            the reimplemented methods.}
    \label{fig:method_compare}
    \vspace{-2mm}
\end{figure}

On Pascal VOC, we conducted an analysis on the influence of the number of shots
on the performance. The results are available in \cref{tab:result_voc}, but to
better catch the trend, these are represented as plots in
\cref{fig:method_compare}. In addition, the same curves are also plotted for
each class individually in \cref{fig:map_per_class}. These two figures highlight
a known trend in few-shot learning: the performance increases with the number of
shots. However, this trend is stronger for novel classes. This is expected as
the models did not receive heavy supervision for these classes. Interestingly,
the trend is less pronounced for novel classes that are really close to some
base classes. As an example, the class \textit{cow} belongs to the novel classes
set but is visually similar to the classes \textit{sheep} and \textit{horse}
from the base set. The mAP for \textit{cow} is high compared to other novel
classes and does not benefit as much from having more shots. The models can
efficiently adapt their representations of the known classes to match the new
one. In \cref{fig:map_per_class}, one can notice decreasing performance when the
number of shots increases for some classes. This is not surprising, support
examples are sampled randomly and can sometimes be poor representatives of their
class, thus misleading the detections. This mostly happens with a small number
of shots as the feature averaging reduces the influence of outliers. As an
example, performance for the class \textit{car} is reduced for all tested
methods between 1 and 3 shots (the support sets are the same for all
methods). The magnitude of this loss depends on the method. FRW, WSAAN and
DANA are more robust than others to poor examples. The opposite can also occur
when the chosen examples are good representatives of their class. As an example,
the class \textit{horse} is suprisingly well detected with 3 shots compared with
1 shot. 

\section{Relative mAP analysis on DOTA, DIOR and Pascal VOC}
\label{app:rmap}

\begin{figure*}[p]
    \centering
    \includegraphics[width=0.8\textwidth]{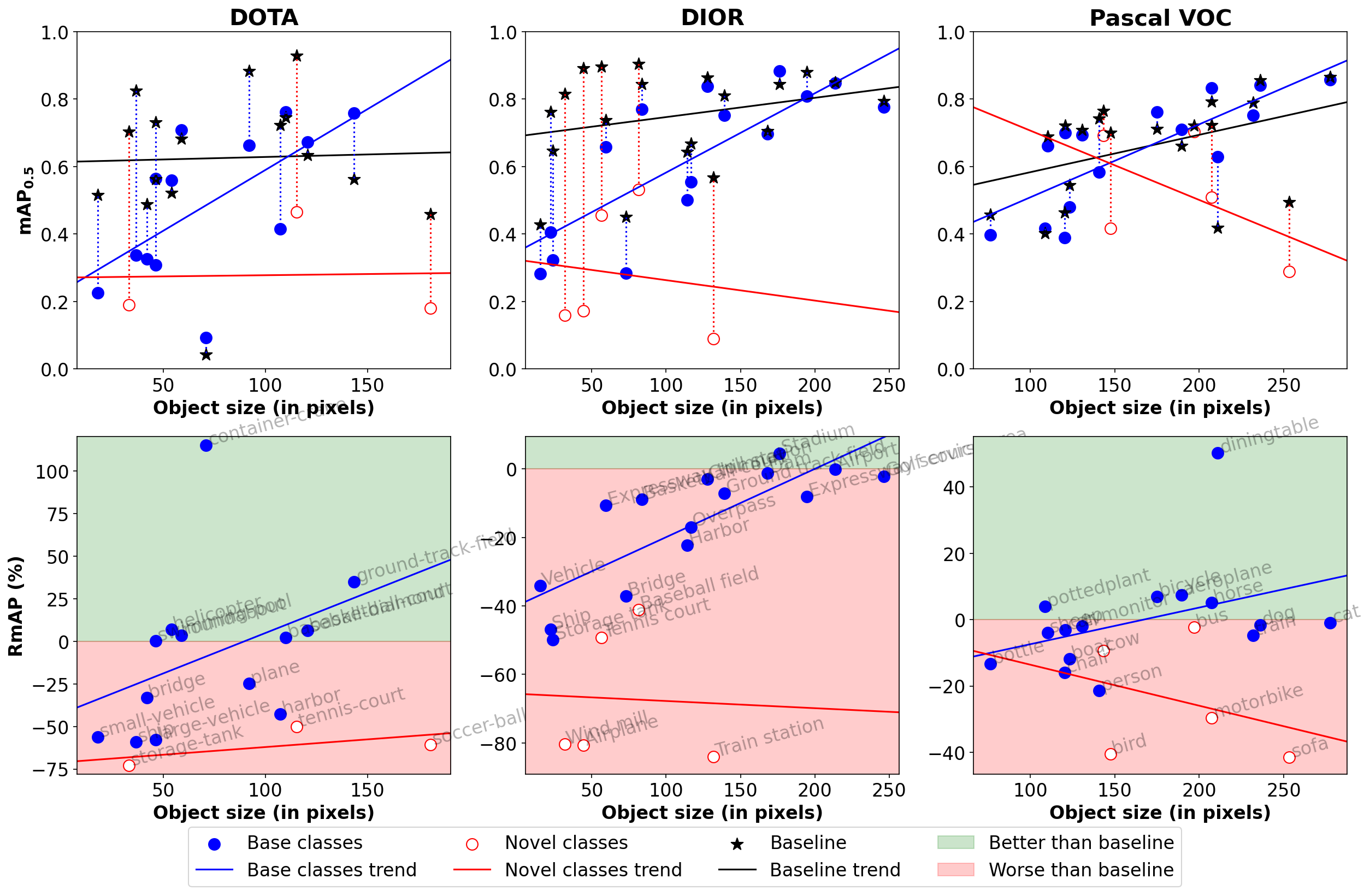}
    \caption{Comparison of FRW (FSOD) and FCOS (baseline) performance against
    object size on each dataset separately: DOTA, DIOR and Pascal VOC. \textbf{(top)}
    absolute $\text{mAP}_{0.5}$ values. \textbf{(bottom)} RmAP computed against the regular
    baseline (i.e. without few-shot).}
    \label{fig:rmap_class}
\end{figure*}

Relative mAP (RmAP), given by \cref{eq:rmap}, is a reliable measure of
how well a FSOD method performs against the classical baseline. However, when
the performance of the baseline is poor, small drops in FSOD regime can lead to
arbitrary large values of RmAP. While, it is still a valid measurement of the
FSOD performance gap, it is not convenient for visualization. One large value
will squeeze the others (mostly contained in the band -80\% , +50\%) and reduce
visibility. This is why we chose to only report the mAP difference in
\cref{fig:perf_by_class}. For the sake of clarity, the same results with RmAP
are visible in \cref{fig:rmap_class}.

\begin{figure*}[p]
    \centering
    \includegraphics[width=0.8\textwidth]{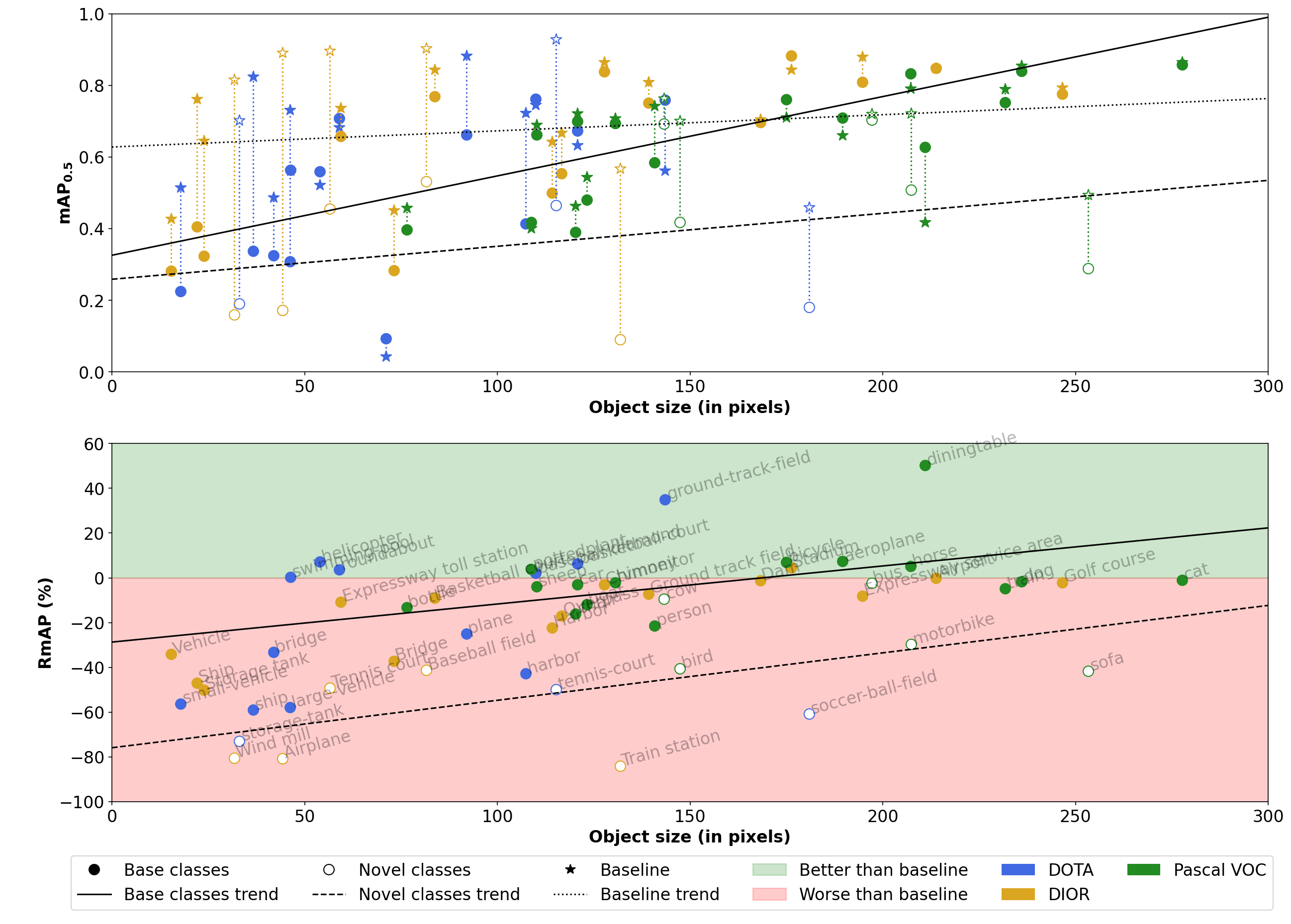}
    \caption{Comparison of FRW (FSOD) and FCOS (baseline) performance against
    object size on the three datasets DOTA, DIOR and Pascal VOC together. \textbf{(top)}
    absolute $\text{mAP}_{0.5}$ values. \textbf{(bottom)} RmAP computed against the regular
    baseline (i.e. without few-shot).}
    \label{fig:rmap_all}
\end{figure*}

We also include in \cref{fig:rmap_all} a plot aggregating the results on the
three datasets of interest. This way, the trend line for novel classes is more
reliable as it is computed from more data points. In addition, it shows that the
trends observed on each dataset are coherent, even between aerial and natural
images. It means that the different performance regimes between aerial and
natural images are explained only by the size of the objects and not by the
different aspects of the objects in the images. Therefore, we could roughly
extrapolate the FSOD performance of a given method on other datasets, only
provided with the objects' size statistics.

\begin{figure}[H]
    \centering
    \includegraphics[width=0.95\columnwidth]{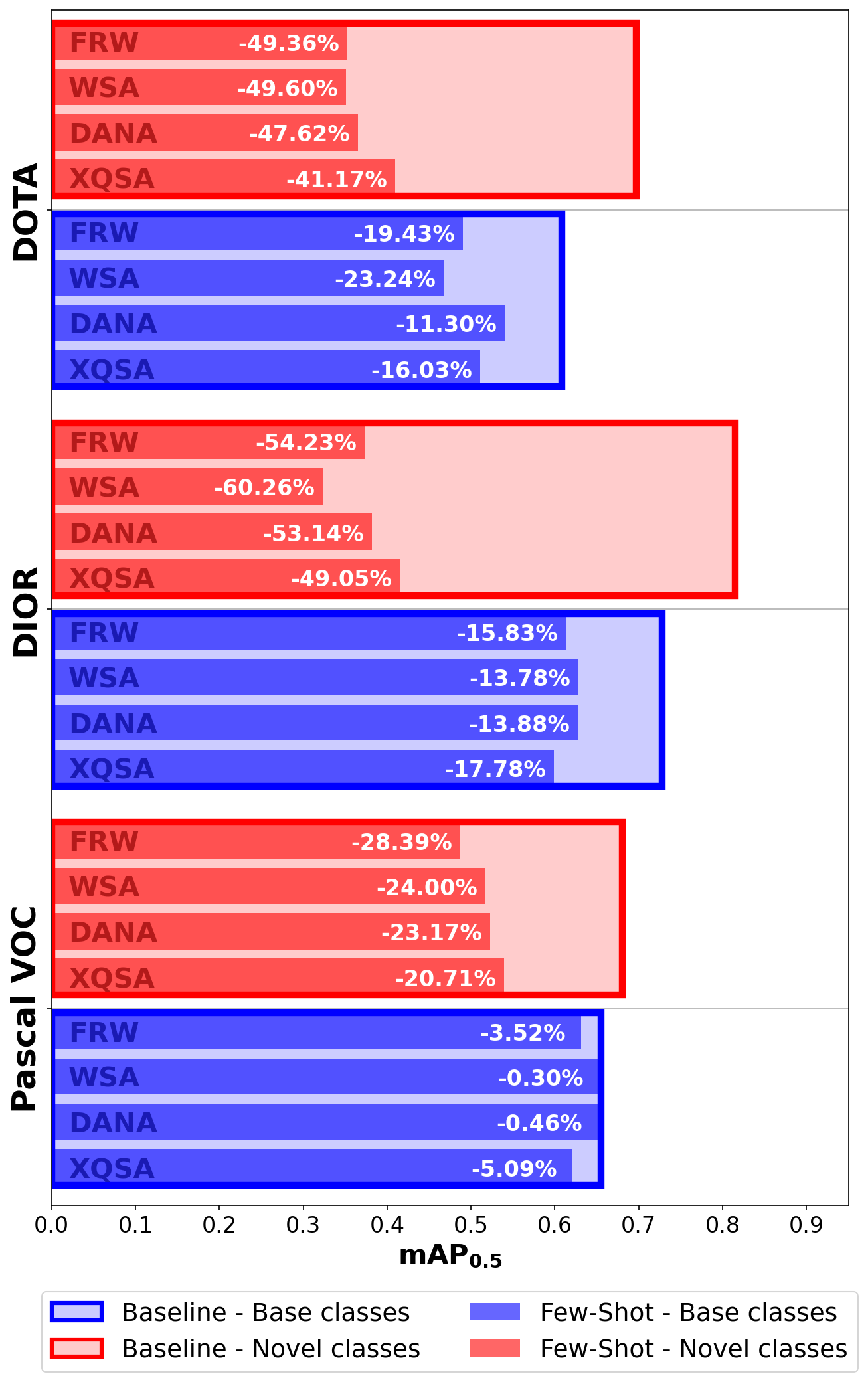}
    \caption{$\text{mAP}_{0.5}$ and corresponding RmAP values of the four best performing
    methods from all our experiments. All methods are trained within our
    proposed AAF framework with data augmentation which explains slightly
    higher performance for FRW and WSA. 10 shots are available for each
    class at inference time.}
    \label{fig:rmap_all_hist}
\end{figure}

In order to assess how well the XQSA method performs on aerial and
natural images, we present in \cref{fig:rmap_all_hist} an extended version of
the plot from \cref{fig:baseline_res} with DANA and XQSA. The exact values of
mAP and RmAP are regrouped in \cref{tab:rmap_values}. This summarizes the
conclusions from \cref{sec:xscale}, XQSA improves largely over the previous
methods on DOTA and DIOR. However, there remains a large performance gap
compared to the no few-shot baseline, i.e. the RmAP (white percentage values on
the bar chart) is still low compared with RmAP on natural images. 

\begin{table}[!h]
    \centering
    \caption{$\text{mAP}_{0.5}$ and RmAP values for some reimplemented methods
    and XQSA with $K=10$ shots. These are the values used to make the plot from
    \cref{fig:rmap_all_hist}}
    \label{tab:rmap_values}
    \resizebox{0.95\columnwidth}{!}{%
    \begin{tabular}{@{}llccccccccl@{}}
    \toprule[1pt]
                                    &               & \multicolumn{2}{c}{\textbf{DOTA}} &  & \multicolumn{2}{c}{\textbf{DIOR}} & \textbf{} & \multicolumn{2}{c}{\textbf{Pascal VOC}} &  \\
                                    &               & \textbf{Base}   & \textbf{Novel}  &  & \textbf{Base}   & \textbf{Novel}  &           & \textbf{Base}      & \textbf{Novel}     &  \\ \cmidrule(r){1-10}
    \multirow{5}{*}{\rotatebox[origin=c]{90}{\textbf{$\text{mAP}_{\mathbf{0.5}}$}}}  & FCOS baseline & 0.61            & 0.70            &  & 0.73            & 0.82            &           & 0.66               & 0.68               &  \\ \cmidrule(lr){2-10}
                                    & FRW           & 0.49            & 0.35            &  & 0.61            & 0.37            &           & 0.63               & 0.49               &  \\
                                    & WSAAN         & 0.47            & 0.35            &  & 0.63            & 0.32            &           & 0.65               & 0.52               &  \\
                                    & DANA          & 0.54            & 0.37            &  & 0.63            & 0.38            &           & 0.65               & 0.52               &  \\
                                    & XQSA          & 0.51            & 0.41            &  & 0.60            & 0.42            &           & 0.51               & 0.47               &  \\
    \multirow{5}{*}{\rotatebox[origin=c]{90}{\textbf{RmAP (\%)}}}                               &               &                 &                 &  &                 &                 &           &                    &                    &  \\
                                    & FRW           & -19.43          & -49.36          &  & -15.83          & -54.23          &           & -3.52              & -28.39             &  \\
                                    & WSAAN         & -23.24          & -49.60          &  & -13.78          & -60.26          &           & -0.30              & -24.00             &  \\
                                    & DANA          & -11.30          & -47.63          &  & -13.88          & -53.14          &           & -0.46              & -23.17             &  \\
                                    & XQSA          & -16.03          & -41.17          &  & -17.78          & -49.05          &           & -21.97             & -31.39             &  \\ \bottomrule[1pt]
    \end{tabular}%
    }
    \end{table}
    
\section{XQSA complementary results}
\label{app:res_coco}

In complement for \cref{tab:xscale_res}, we also provide the comparison between
XQSA, DANA and FRW with $\text{mAP}_{0.5:0.95}$ metric and on MS COCO datasets.
The results are provided in
\cref{tab:res_comp_dota,tab:res_comp_dior,tab:res_comp_pascal,tab:xqsa_coco_res,tab:res_comp_coco}.
Results on MS COCO are in agreement with 
the results on Pascal VOC. XQSA is
beneficial especially for novel and small objects. Overall, as MS COCO mostly
contains medium and large objects, the performance of XQSA is just barely better
than DANA's and FRW's on novel classes. However, on base classes, the
performance drops significantly.

$\text{mAP}_{0.5:0.95}$ is a more demanding metric for object detection. It is
especially hard for small objects as a few pixels shift from ground truth can
greatly reduce the IoU and therefore lead to a missed detection. This
intensifies as the IoU threshold increases in the mAP computation. For DOTA and
DIOR, results with $\text{mAP}_{0.5:0.95}$ are in agreement with results from
\cref{tab:xscale_res} (i.e. with $\text{mAP}_{0.5}$). However, XQSA does not
perform better than DANA on Pascal VOC and MS COCO novel classes with
$\text{mAP}_{0.5:0.95}$. This is mainly due to the metric being too strict on
small objects. This questions the soundness of these metrics for FSOD,
especially when dealing with small objects.

    \begin{table}[!h]
        \centering
        \caption{Performance comparison between FRW, DANA and XQSA (ours) on
        DOTA. Reported values are $\text{mAP}_{0.5:0.95}$.}
        \label{tab:res_comp_dota}
        \resizebox{\columnwidth}{!}{%
        \begin{tabular}{@{}lcrrrrrrrr@{}}
            \toprule[1pt]
            \multicolumn{1}{c}{}          & \multicolumn{9}{c}{$\text{mAP}_{0.5:0.95}$ \textbf{ on DOTA}}                                                                                                                                                                                                                                                         \\ \cmidrule{2-10} 
            \multicolumn{1}{c}{\textbf{}} & \multicolumn{4}{c}{\textbf{Base}}                                                                                            & \multicolumn{1}{l}{} & \multicolumn{4}{c}{\textbf{Novel}}                                                                                                  \\ \midrule
            \multicolumn{1}{c}{\textbf{}} & \textbf{All}                       & \multicolumn{1}{c}{\textbf{S}} & \multicolumn{1}{c}{\textbf{M}} & \multicolumn{1}{c}{\textbf{L}} & \multicolumn{1}{l}{} & \multicolumn{1}{c}{\textbf{All}} & \multicolumn{1}{c}{\textbf{S}} & \multicolumn{1}{c}{\textbf{M}} & \multicolumn{1}{c}{\textbf{L}} \\
            \textbf{FRW}                  & \multicolumn{1}{r}{0.232}          & 0.086                          & 0.278                          & 0.322                          &                      & 0.160                            & 0.042                          & 0.141                          & 0.296                          \\
            \textbf{DANA}                 & \multicolumn{1}{r}{\textbf{0.266}} & \textbf{0.114}                 & \textbf{0.307}                 & \textbf{0.376}                 &                      & 0.172                            & 0.056                          & 0.204                          & \textbf{0.324}                 \\
            \textbf{XQSA}                 & \multicolumn{1}{r}{0.253}          & 0.089                          & 0.288                          & 0.346                          &                      & \textbf{0.210}                   & \textbf{0.079}                 & \textbf{0.252}                 & 0.265                          \\ \bottomrule[1pt]
            \end{tabular}%
        }
        \end{table}

    \begin{table}[!h]
        \centering
        \caption{Performance comparison between FRW, DANA and XQSA (ours) on
        DIOR. Reported values are $\text{mAP}_{0.5:0.95}$.}
        \label{tab:res_comp_dior}
        \resizebox{\columnwidth}{!}{%
        \begin{tabular}{@{}lcrrrrrrrr@{}}
            \toprule[1pt]
            \multicolumn{1}{c}{}          & \multicolumn{9}{c}{$\text{mAP}_{0.5:0.95}$ \textbf{ on DIOR}}                                                                                                                                                                                                                                                         \\ \cmidrule{2-10}  
            \multicolumn{1}{c}{\textbf{}} & \multicolumn{4}{c}{\textbf{Base}}                                                                                            & \multicolumn{1}{l}{} & \multicolumn{4}{c}{\textbf{Novel}}                                                                                                  \\ \midrule
            \multicolumn{1}{c}{\textbf{}} & \textbf{All}                       & \multicolumn{1}{c}{\textbf{S}} & \multicolumn{1}{c}{\textbf{M}} & \multicolumn{1}{c}{\textbf{L}} & \multicolumn{1}{l}{} & \multicolumn{1}{c}{\textbf{All}} & \multicolumn{1}{c}{\textbf{S}} & \multicolumn{1}{c}{\textbf{M}} & \multicolumn{1}{c}{\textbf{L}} \\
            \textbf{FRW}                  & \multicolumn{1}{r}{0.356}          & 0.026                          & 0.230                          & 0.508                          &                      & 0.200                            & 0.005                          & 0.170                          & 0.333                          \\
            \textbf{DANA}                 & \multicolumn{1}{r}{\textbf{0.364}} & \textbf{0.035}                 & \textbf{0.249}                 & \textbf{0.523}                 &                      & 0.204                            & 0.008                          & 0.175                          & 0.340                          \\
            \textbf{XQSA}                 & \multicolumn{1}{r}{0.348}          & \textbf{0.035}                 & 0.229                          & 0.515                          &                      & \textbf{0.228}                   & \textbf{0.010}                 & \textbf{0.210}                 & \textbf{0.348}                 \\ \bottomrule[1pt]
            \end{tabular}%
        }
        \end{table}

    \begin{table}[!h]
        \centering
        \caption{Performance comparison between FRW, DANA and XQSA (ours) on
        Pascal VOC. Reported values are $\text{mAP}_{0.5:0.95}$.}
        \label{tab:res_comp_pascal}
        \resizebox{\columnwidth}{!}{%
        \begin{tabular}{@{}lcrrrrrrrr@{}}
            \toprule[1pt]
            \multicolumn{1}{c}{}          & \multicolumn{9}{c}{$\text{mAP}_{0.5:0.95}$ \textbf{ on Pascal VOC}}                                                                                                                                                                                                                               \\ \cmidrule{2-10} 
            \multicolumn{1}{c}{\textbf{}} & \multicolumn{4}{c}{\textbf{Base}}                                                                                            &  & \multicolumn{4}{c}{\textbf{Novel}}                                                                                                  \\ \midrule
            \multicolumn{1}{c}{\textbf{}} & \textbf{All}              & \multicolumn{1}{c}{\textbf{S}} & \multicolumn{1}{c}{\textbf{M}} & \multicolumn{1}{c}{\textbf{L}} &  & \multicolumn{1}{c}{\textbf{All}} & \multicolumn{1}{c}{\textbf{S}} & \multicolumn{1}{c}{\textbf{M}} & \multicolumn{1}{c}{\textbf{L}} \\
            \textbf{FRW}                  & \multicolumn{1}{r}{0.379} & 0.065                          & 0.228                          & 0.505                          &  & 0.291                            & 0.056                          & 0.122                          & 0.400                          \\
            \textbf{DANA}                 & \multicolumn{1}{r}{\textbf{0.391}} & \textbf{0.127}                 & \textbf{0.064}                 & \textbf{0.317}                 &  & \textbf{0.317 }                  & 0.052                          & 0.111                          & \textbf{0.434}                          \\
            \textbf{XQSA}                 & \multicolumn{1}{r}{0.274} & 0.032                          & 0.166                          & 0.368                          &  & 0.251                            & \textbf{0.064}                 & \textbf{0.127}                 & 0.352                          \\ \bottomrule[1pt]
            \end{tabular}%
        }
        \end{table}

    \begin{table}[!h]
        \centering
        \caption{Performance comparison between FRW, DANA and XQSA (ours) on
        MS COCO. Reported values are $\text{mAP}_{0.5}$.}
        \label{tab:xqsa_coco_res}
        \resizebox{\columnwidth}{!}{%
        \begin{tabular}{@{}lcrrrrrrrr@{}}
            \toprule[1pt]
            \multicolumn{1}{c}{}          & \multicolumn{9}{c}{$\text{mAP}_{0.5}$ \textbf{ on MS COCO}}                                                                                                                                                                                                                                                      \\ \cmidrule{2-10} 
            \multicolumn{1}{c}{\textbf{}} & \multicolumn{4}{c}{\textbf{Base}}                                                                                                     &                      & \multicolumn{4}{c}{\textbf{Novel}}                                                                                                  \\ \midrule
            \multicolumn{1}{c}{\textbf{}} & \textbf{All}                       & \multicolumn{1}{c}{\textbf{S}} & \multicolumn{1}{c}{\textbf{M}} & \multicolumn{1}{c}{\textbf{L}} &                      & \multicolumn{1}{c}{\textbf{All}} & \multicolumn{1}{c}{\textbf{S}} & \multicolumn{1}{c}{\textbf{M}} & \multicolumn{1}{c}{\textbf{L}} \\
            \textbf{FRW}                  & \multicolumn{1}{r}{0.290}          & 0.131                          & 0.359                          & 0.480                          &                      & 0.241                            & 0.115                          & 0.225                          & \textbf{0.387}                          \\
            \textbf{DANA}                 & \multicolumn{1}{r}{\textbf{0.381}} & \textbf{0.233}                 & \textbf{0.518}                 & \textbf{0.564}                 & \multicolumn{1}{r}{} & 0.247                            & 0.120                          & \textbf{0.294 }                & 0.379                          \\
            \textbf{XQSA}                 & \multicolumn{1}{r}{0.316}          & 0.161                          & 0.401                          & 0.498                          & \multicolumn{1}{r}{} & \textbf{0.250}                   & \textbf{0.126}                 & 0.261                          & 0.385                          \\ \bottomrule[1pt]
            \end{tabular}
        }
        \end{table}

    \begin{table}[H]
        \centering
        \caption{Performance comparison between FRW, DANA and XQSA (ours) on
        MS COCO. Reported values are $\text{mAP}_{0.5:0.95}$.}
        \label{tab:res_comp_coco}
        \resizebox{\columnwidth}{!}{%
        \begin{tabular}{@{}lcrrrrrrrr@{}}
            \toprule[1pt]
            \multicolumn{1}{c}{}          & \multicolumn{9}{c}{$\text{mAP}_{0.5:0.95}$ \textbf{ on MS COCO}}                                                                                                                                                                                                                                                      \\ \cmidrule{2-10} 
            \multicolumn{1}{c}{\textbf{}} & \multicolumn{4}{c}{\textbf{Base}}                                                                                            & \multicolumn{1}{l}{} & \multicolumn{4}{c}{\textbf{Novel}}                                                                                                  \\ \midrule
            \multicolumn{1}{c}{\textbf{}} & \textbf{All}                       & \multicolumn{1}{c}{\textbf{S}} & \multicolumn{1}{c}{\textbf{M}} & \multicolumn{1}{c}{\textbf{L}} & \multicolumn{1}{l}{} & \multicolumn{1}{c}{\textbf{All}} & \multicolumn{1}{c}{\textbf{S}} & \multicolumn{1}{c}{\textbf{M}} & \multicolumn{1}{c}{\textbf{L}} \\
            \textbf{FRW}                  & \multicolumn{1}{r}{0.156}          & 0.055                          & 0.188                          & 0.278                          &                      & 0.124                            & 0.048                          & 0.109                          & 0.208                          \\
            \textbf{DANA}                 & \multicolumn{1}{r}{\textbf{0.225}} & \textbf{0.102}                 & \textbf{0.297}                 & \textbf{0.365}                 &                      & \textbf{0.134}                   & \textbf{0.053}                 & \textbf{0.150}                 & \textbf{0.215}                 \\
            \textbf{XQSA}                 & \multicolumn{1}{r}{0.114}          & 0.044                          & 0.142                          & 0.320                          &                      & 0.103                            & 0.049                          & 0.100                          & 0.167                          \\ \bottomrule[1pt]
            \end{tabular}%
        }
        \end{table}

\section{XQSA Ablation study}
\label{app:ablation}

To confirm the benefits of each component of our attention methods, we conduct
a brief ablation experiment, adding separately the different modules of our
proposed attention mechanism. The ablation is conducted on DOTA and the results
are available in \cref{tab:xqsa_ablation}. From this table, it is clear that
each component plays a role in the better performance of our method. Both the
fusion (with the MLP) and the skip connections around fusion and alignment are
beneficial for the performance on novel classes. It is worth to note that
Background Attenuation proposed by \cite{chen2021should} helps both for base
and novel classes, which confirms the experiments conducted by the authors.

\begin{table}[!h]
    \centering
    \caption{Ablation study of the XQSA attention method on DOTA dataset. Each
    component is added separately to assess its respective benefits for
    detection. mAP scores are reported for base and novel classes with $K=10$
    shots.}
    \label{tab:xqsa_ablation}
    \resizebox{0.9\columnwidth}{!}{%
    \begin{tabular}{@{}llcccc@{}}
    \toprule[1pt]
    Multiscale Alignment      &  & \checkmark                & \checkmark                & \checkmark                & \checkmark                         \\
    MLP Fusion                &  &                           & \checkmark                & \checkmark                & \checkmark                         \\
    Skip Connections          &  &                           &                           & \checkmark                & \checkmark                         \\
    Background Attenuation    &  &                           &                           &                           & \checkmark                         \\ \midrule
    \textbf{Base classes}     &  & \multicolumn{1}{r}{0.492} & \multicolumn{1}{r}{0.495} & \multicolumn{1}{r}{0.491} & \multicolumn{1}{r}{\textbf{0.511}} \\
    \textbf{Novel classes}    &  & \multicolumn{1}{r}{0.365} & \multicolumn{1}{r}{0.388} & \multicolumn{1}{r}{0.403} & \multicolumn{1}{r}{\textbf{0.410}} \\ \bottomrule[1pt]
    \end{tabular}%
    }
    \end{table}

\end{document}